\documentclass[10pt,letterpaper]{article}
\usepackage{multirow}
\usepackage{ccn}
\usepackage{pslatex}
\usepackage{apacite}
\usepackage{hyperref}
\usepackage{natbib}
\usepackage{lineno}
\usepackage{graphicx}
\usepackage{subfigure}
\usepackage{enumitem}
\usepackage{amsfonts}
\usepackage{amsmath, amssymb, bm}

\title{Modeling the Human Visual System: Comparative Insights from Response-Optimized and Task-Optimized Vision Models, Language Models, and Different Readout Mechanisms}
 
\author{{\large \bf Shreya Saha (ssaha@ucsd.edu)} \\
Electrical and Computer Engineering \\
University of California, San Diego
  \AND {\large \bf Ishaan Chadha (ichadha@ucsd.edu)} \\
Hal{\i}cıoğlu Data Science Institute \\
University of California, San Diego
\AND {\large \bf Meenakshi Khosla (mkhosla@ucsd.edu)} \\
Department of Cognitive Science, Department of Computer Science and Engineering \\
University of California, San Diego}

\begin{document}

\maketitle

\section{Abstract}
{
\bf
Over the past decade, predictive modeling of neural responses in the primate visual system has advanced significantly, driven by diverse deep neural network approaches. These include models optimized for visual recognition, methods that align visual and language information, models trained directly on brain data, and representations from large language models (LLMs). Additionally, various readout mechanisms have been developed to map network activations to neural responses. Despite this progress, it remains unclear which approach performs best across different regions of the visual hierarchy.
In this study, we systematically compare these methods for modeling the human visual system and propose novel strategies to enhance response predictions. We demonstrate that the choice of readout mechanism significantly impacts prediction accuracy and introduce a biologically grounded readout that dynamically adjusts receptive fields based on image content and learns geometric invariances of voxel responses directly from data. This novel readout outperforms factorized methods by 3-23$\%$ and standard ridge regression by 7-53$\%$, setting a new benchmark for neural response prediction.
Our findings reveal distinct modeling advantages across the visual hierarchy: response-optimized models with visual inputs excel in early to mid-level visual areas, while embeddings from LLMs—leveraging detailed contextual descriptions of images—and task-optimized models pretrained on large vision datasets provide the best fit for higher visual regions. Through comparative analysis, we identify three functionally distinct regions in the visual cortex: one sensitive to perceptual features not captured by linguistic descriptions, another attuned to fine-grained visual details encoding semantic information, and a third responsive to abstract, global meanings aligned with linguistic content. Together, these findings offer key insights into building more precise models of the visual system.	

}
\begin{quote}
\small
\textbf{Keywords:} 
Neuro AI, vision, deep neural networks, Neural Response Modeling, fMRI encoding, Readout Mechanisms, Vision Language Alignment
\end{quote}

\begin{figure*}[h!]
    \centering
    \includegraphics[width=0.81\linewidth,trim={0 0 0 0},clip]{figures/Fig1.pdf}
    \caption{(A) High-level schematic of the key components analyzed in this study. (B) Various stimuli used to model the visual cortex. (C) Different encoder backbones employed in the study. (D) Readout mechanisms (Linear, Gaussian, Factorized, and Semantic Spatial Transformer) that map ANN encoder representations to neuronal or voxel responses.}
    \label{fig:intro}
\end{figure*}

\section{Introduction and Related Work}

Building accurate predictive models of the visual system has been a longstanding goal in neuroscience. Early approaches primarily relied on handcrafted features, such as Gabor filters, curvature models, and motion energy models, to predict responses in early to mid-level visual areas~\citep{hubel1962receptive, livingstone1984anatomy,albrecht1982striate, gallant1993selectivity,hubel1968receptive,desimone1984stimulus,tanaka1991coding,pasupathy2002population,yue2020curvature,yang2023unimo,pasupathy1999responses,tsunoda2001complex,rust2010selectivity,brincat2004underlying,zeki1973colour,pasupathy2001shape,moran1985selective,kobatake1994neuronal,kriegeskorte2008matching,kobatake1998effects,miyashita1988neuronal}. Similarly, word-based descriptions were often used to model responses in higher-level visual regions~\citep{huth2012continuous}. These models provided interpretability, as the features they employed were well understood and linked to specific visual computations. However, they lacked quantitative precision in their ability to predict neural responses. \footnote{Code can be found at - \href{https://github.com/NeuroML-Lab/Visual-Stream-Modeling}{https://github.com/NeuroML-Lab/Visual-Stream-Modeling}}

The advent of deep convolutional neural networks (DCNNs) marked a significant improvement in predictive accuracy across the visual system ~\citep{yamins2014performance,abdelhack2018sharpening,wen2018neural,horikawa2017generic,eickenberg2017seeing,gucclu2015deep,cichy2016comparison,khaligh2014deep, schrimpf2020integrative, storrs2021diverse,safarani2021towards,schwartz2019inducing,seeliger2021end, shenhigh}. DCNNs trained on image categorization tasks emerged as the first class of models capable of capturing neural activity in the primate visual cortex with a reasonable degree of fidelity. This success spurred a wave of model-brain comparisons, wherein variations in input data, architecture, and learning objectives were explored to identify the most predictive models of brain responses in both non-human primates and humans.

More recently, models trained using multimodal contrastive learning approaches, such as CLIP, or image-caption embeddings from large language models (LLMs), have shown promise in predicting neural responses in the visual cortex \citep{tang2024brain, wang2022incorporating, doerig2024Visual}. These findings suggest that visual brain responses may encode some linguistically learned structure or semantics. In parallel, another class of models, optimized specifically for neural response prediction \citep{khosla2022high, khosla2022characterizing,federer2020improved,dapello2022aligning,st2023brain} — either trained from scratch or fine-tuned to better align with primate visual representations—has achieved impressive predictive accuracy, particularly with the availability of large-scale neural datasets~\citep{allen2022massive}.

Given the broad range of modeling approaches applied to different regions of the visual cortex, a critical question remains: which approach offers the most quantitatively precise predictions of neural responses across the various areas of the human visual system? This challenge underscores the need for systematic comparisons to determine the optimal models for different visual processing stages. While some recent studies have made strides in conducting large-scale comparative analyses, they tend to focus primarily on specific pre-selected visual regions and largely compare different task-optimized vision networks~\citep{conwell2022can}. A more comprehensive comparison is needed to evaluate a broader set of approaches, including models based on response optimization and embeddings from language models trained on vision-aligned tasks or pure language data.

An equally pressing issue concerns the readout mechanism by which models’ internal representations are mapped onto neural responses \cite{ivanova2022beyond}. The predominant readout in primate studies is the fully-connected affine readout, often used in regularized linear regression models. However, these linear ridge regression readouts require numerous parameters, especially in high-dimensional spaces, leading to significant computational and memory demands. To mitigate this, more efficient methods have been developed, such as factorized linear readouts by \citep{klindt2017neural}, that decouple spatial from feature selectivity, reducing overhead and improving prediction accuracy. The Gaussian2D readout \citep{lurz2020generalization} further enhances parameter efficiency by learning spatial readout locations using a bivariate Gaussian distribution informed by anatomical retinotopy. However, it is still unclear which readout approach provides the best predictive accuracy across different cortical areas.

Determining the most effective model—and the most suitable readout—for each region of the visual cortex is vital. Accurate models provide a powerful platform for \emph{in silico} experimentation, enabling researchers to test hypotheses that may be impractical to probe \emph{in vivo}. They also inform experimental design and facilitate precise neural population control \citep{walker2019inception, bashivan2019neural}. In this way, achieving high predictive accuracy is foundational for both practical applications and deeper theoretical insights into visual processing.

In this paper, we bridge these gaps by systematically comparing a broad array of models—along with diverse readout methods—to identify the most accurate approach for each region of the human visual cortex. Specifically, we make the following key contributions:
\begin{enumerate}[noitemsep]
\item \textbf{Comprehensive analysis of different neural network models and readouts:} We systematically compare an extensive set of neural network models spanning vision-only, vision-language and language-only paradigms. Additionally, we explore different readout mechanisms and examine which models perform better in specific brain regions, while highlighting the unique advantages each provides.

    \item \textbf{Introduction of a novel readout:} We introduce a novel biologically-grounded readout method which delivers significant improvements in accuracy, outperforming factorized methods by 3-23$\%$ and standard ridge regression (the de facto choice in many studies)  by 7-53$\%$ .
    
    \item \textbf{Identification of brain regions sensitive to perceptual and semantic information:} Through large-scale comparative analysis of models across various visual regions, we identify three distinct regions in the human visual cortex that respond primarily to (a) low-level perceptual characteristics of the input, (b) localized visual semantics aligned with linguistic descriptions, and (c) global semantic interpretations of the input, also aligned with language.

\end{enumerate}

\section{Methods}

\subsection{Encoders}

\subsubsection{Task-optimized Models} We use encoders from pre-trained models like AlexNet~\citep{krizhevsky2012imagenet} and ResNet~\citep{he2016deep}, originally trained for object classification on the large-scale ImageNet dataset~\citep{deng2009imagenet}. The weights of their intermediate layers are frozen, and only the readout layers (described later) are trained. Prior research shows that early layers of neural networks align with lower visual cortex regions, while later layers correspond to higher regions \citep{khaligh2014deep, gucclu2015deep, cichy2016comparison, eickenberg2017seeing, horikawa2017generic, wen2018neural, abdelhack2018sharpening, yamins2014performance}. Thus, we experimented with all layers of task-optimized networks. For fair comparison, we selected the best-performing layers for each cortical region (see Appendix Table \ref{tab:Task_Optimized_model_comp} and summary in Table \ref{tab:Results_summary}). 

\subsubsection{Response-optimized Models} Task-optimized models often rely heavily on a priori hypotheses, which may be biased towards pre-existing conclusions, limiting novel discoveries. Further, these networks are typically optimized for specific tasks, such as object classification, which may not capture the full range of visual processing in the cortex. Recently, \citep{khosla2022high} showed that training neural networks from scratch with stimulus images and fMRI data from the NSD dataset \citep{allen2022massive} can achieve accuracy comparable to state-of-the-art task-optimized models. By directly optimizing for neural responses, these models are free to learn representations that are more closely aligned with the underlying neural computations, unencumbered by the biases inherent in task-driven models. This flexibility can enable response-optimized models to uncover richer, more generalizable representations that better reflect the diversity of neural activation patterns across brain regions. 

We leverage the same architecture for response-optimized models as prior work~\citep{khosla2022high}, which consists of a convolutional neural network (CNN) core that transforms raw input data into feature spaces characteristic of different brain regions, followed by a readout layer that maps these features to fMRI voxel responses. The core contains four convolutional blocks, where each convolutional block includes two convolutional layers, followed by internal batch normalization, nonlinear ReLU activations, and an anti-aliased average pooling operation. To ensure equivariance under all isometries, we use E(2)-Equivariant Steerable Convolution layers \citep{weiler2019general}. Further analysis on the importance of network architecture for Response-optimized models can be found in Appendix section \nameref{subsec:task_response_architecture} and Table \ref{tab:task_response_architecture}.

\subsubsection{Language Models} - Recent studies show that higher visual regions converge toward representational formats similar to large language model (LLM) embeddings of scene descriptions. \citep{doerig2024Visual} used MPNET \citep{song2020mpnet} to encode image captions and map them to fMRI responses via ridge regression, finding it effectively modeled higher visual areas despite being trained on language inputs alone. In contrast, \citep{tang2024brain} and \citep{wang2022incorporating} used multimodal models like CLIP \citep{radford2021learning} and BridgeTower \citep{yang2023unimo}, showing that CLIP outperforms vision-only models in capturing higher visual regions, attributing this to language feedback. These motivated us to evaluate language models relative to vision-only response-optimized and task-optimized models as detailed below (More detailed comparison on CLIP and MPNET embeddings and additional results with GPT2-XL \cite{brown2020language} can be found in Appendix section \nameref{subsec:cm} and Table \ref{tab:Response_Optimized_lANGUAGE_model_comp}) -

\begin{enumerate}[noitemsep]
        \item \textbf{Single Caption} - Images in the NSD dataset are sourced from MS COCO \citep{lin2014microsoft} and annotated by 4-5 human annotators. We encode these captions using CLIP or MPNET, average the encodings, and input them into a linear regressor to map them to fMRI voxel responses. Since the captions describe the image as a whole without offering spatial details (i.e., fine-grained delineations of features at different locations), we only use the ridge linear readout for single caption inputs.

        \item \textbf{Dense Caption} - An image of size $424*424$ is divided into grids of size $53*53$. For each grid, a caption is generated using GPT-2, which is then encoded by either CLIP or MPNET. Thus an image of shape $3*424*424$ is transformed into a feature representation $N*8*8$, where N is the size of the embedding produced by CLIP or MPNET. The dense-caption language encoders further process these feature maps through a single convolutional block (as described earlier for the response-optimized vision encoders) before passing them to the readout model. For additional technical details—including an in-depth examination of whether dense-caption improvements stem from spatial subdivision or increased semantic detail, as well as experiments comparing alternative single-caption approaches—please refer to the Appendix section \nameref{subsec:dense_captions}, Table \ref{tab:ssingle_caption_analysis} and Figure \ref{fig:densified_single_captions}.
\end{enumerate}

\subsection{Readouts}
The encoders discussed above are paired with a readout model (Figure \ref{fig:intro}) that maps the encoder feature representations to voxel fMRI responses from various regions of the visual cortex.

\subsubsection{Linear Readout}
This approach uses a ridge regression model to map encoder features directly to voxel responses. Let \(n\) be the total number of voxels in the measured brain region. For a given stimulus \(i\), the predicted voxel response vector \(\hat{\mathbf{Y}}_i \in \mathbb{R}^n\) is computed as $\hat{\mathbf{Y}}_i = \mathbf{W}\,\mathbf{E}_i$, where \(\mathbf{E}_i \in \mathbb{R}^e\) is the flattened encoder feature representation and \(\mathbf{W} \in \mathbb{R}^{n \times e}\) is the weight matrix. These weights are learned by minimizing the ridge regression objective: $\min_{\mathbf{W}} \|\mathbf{Y} - \mathbf{W}\,\mathbf{E}\|_F^2 + \lambda \|\mathbf{W}\|_F^2$, where \(\mathbf{Y}\) is the matrix of true voxel responses, \(\mathbf{E}\) is the corresponding matrix of encoder features, \(\|\cdot\|_F\) denotes the Frobenius norm, and \(\lambda\) is the regularization parameter. We select the optimal \(\lambda\) via cross-validation.



\subsubsection{Spatial-Feature Factorized Linear Readout} factorizes the linear readout model into spatial (the portion of the input space a voxel is sensitive to) and feature (the specific features of the input space a voxel responds to) dimensions, as described in \citep{klindt2017neural}. By separating spatial (where) and feature (what) dimensions, the model mirrors the known structure of neural receptive fields in the brain, where neurons exhibit sensitivity to specific spatial locations and particular feature types. This approach not only significantly reduces the number of parameters but also aligns more closely with the known characteristics of neural responses. 
\begin{equation}
    \hat{Y}_{c,n} 
    = \sum_{w=1}^{W} \sum_{h=1}^{H} \mathbf{E}_{c,w,h} \, \mathbf{S}_{n,w,h},
    \quad
    \hat{Y}_n 
    = \sum_{c=1}^{C} \hat{Y}_{c,n} \, \mathbf{F}_{n,c}.
    \label{eq:spatial_linear1}
\end{equation}


Here, \(\hat{Y}_n\) represents the predicted response for voxel \(n\), and \(\mathbf{E} \in \mathbb{R}^{C \times W \times H}\) is the encoder feature map (the ``what''). The spatial weights \(\mathbf{S} \in \mathbb{R}^{N \times W \times H}\) specify the receptive field (the ``where'') for each of the \(N\) voxels, while the feature weights \(\mathbf{F} \in \mathbb{R}^{N \times C}\) determine each voxel’s sensitivity to the \(C\) feature channels. \(W\) and \(H\) denote the spatial dimensions of the encoder feature map.
\\

\subsubsection{Gaussian 2D Readout} This readout models each voxel's spatial sensitivity as a 2D Gaussian in the encoder feature space~\citep{lurz2020generalization}. Specifically, each voxel \(n\) is associated with a bivariate Gaussian distribution \(G_n(x,y) \sim \mathcal{N}(\mu_n, \Sigma_n)\), whose mean \(\mu_n\) represents the voxel's preferred location (receptive field center), and whose covariance \(\Sigma_n\) defines the size, shape, and orientation of the receptive field along the \(x\) and \(y\) axes. The same spatial Gaussian is applied uniformly across all feature channels, indicating a shared positional sensitivity for each voxel.\\
To compute the response \(\hat{Y}_n\) of voxel \(n\), we first bilinearly interpolate the feature values \(\mathbf{V}_c(x, y)\) from channel \(c\) of the encoder feature map \(\mathbf{E} \in \mathbb{R}^{C \times W \times H}\) at spatial coordinates \((x, y)\), weighted by the Gaussian distribution \(G_n(x,y)\). We then multiply these interpolated values by the learned channel-specific weights \(\mathbf{W}_{nc}\) and sum over channels: $\hat{Y}_n = \sum_{c=1}^{C} \mathbf{W}_{nc} \,\mathbf{V}_c(x, y).$
Here, \(\mathbf{W}_{nc}\) determines the contribution of channel \(c\) to voxel \(n\), and the interpolated feature \(\mathbf{V}_c(x,y)\) depends on the Gaussian weighting specified by \(G_n(x,y)\). By incorporating spatial information in this way, the Gaussian 2D readout captures the spatial sensitivity of each voxel with fewer parameters than the Spatial-Feature Factorized Linear Readout.
\\

\subsubsection{Semantic Spatial Transformer Readout} 
\label{subsubsec:stn_readout}

\begin{figure*}[h!]
    \centering
    \includegraphics[width=1.0\linewidth,trim={0 0 0 0},clip]{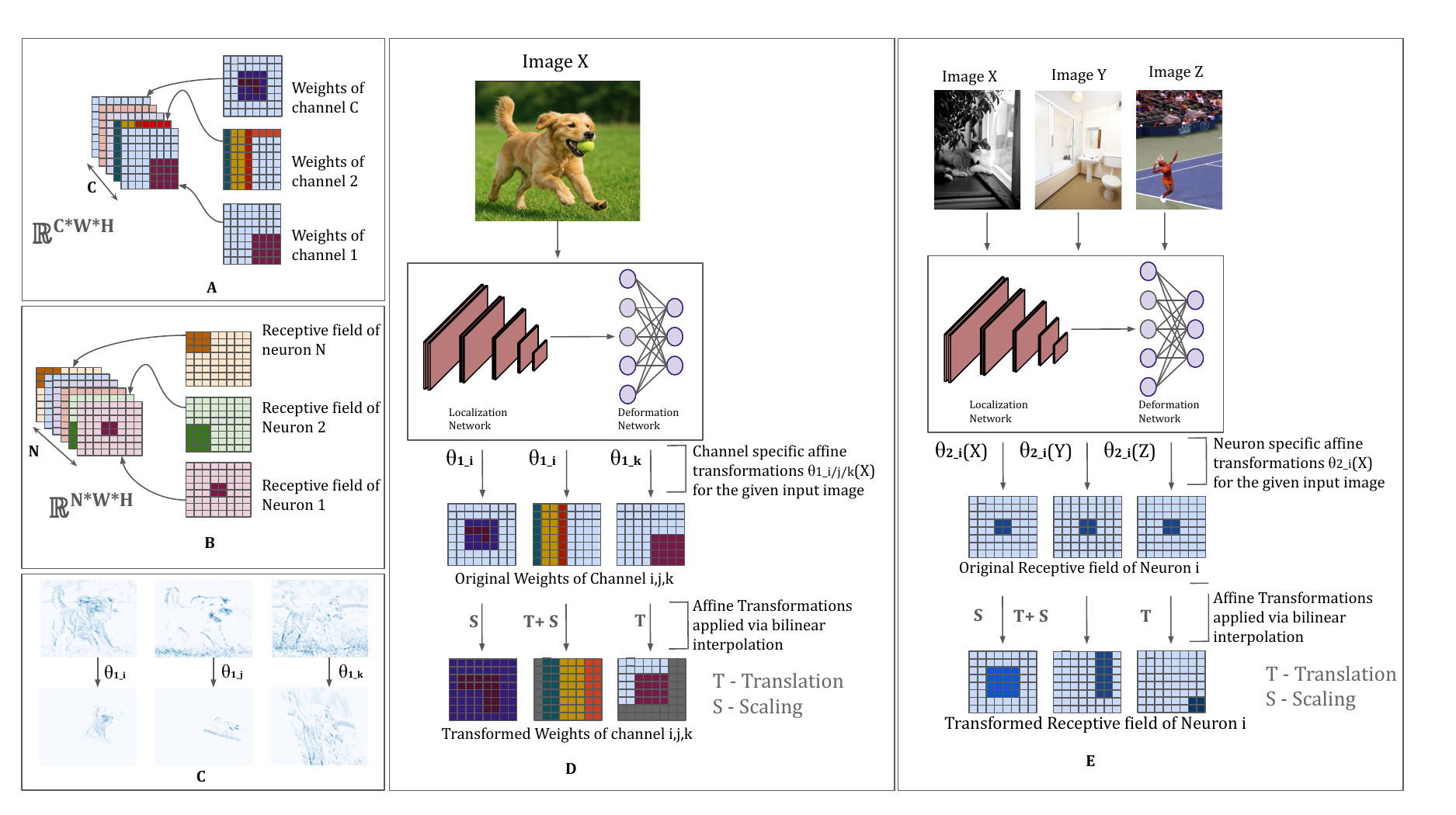}
    \caption{\textbf{Semantic Spatial Transformer Readout} (A) Schematic of encoder feature maps; color intensity reflects weight magnitude for interpretability. (B) Schematic of the learned Spatial Weights (“Where”) matrix, which defines the receptive fields of the $N$ modeled neurons. Each neuron's receptive field is shown in a distinct color, with darker intensities highlighting its spatial extent (shape and location).
    (C–D) Input-dependent modulation of feature maps. (C) Example affine transformations applied to feature maps in response to the input image shown in (D). Top row: original feature maps; bottom row: corresponding transformed maps after applying learned affine transformations. Affine spatial transformations serve to reformat the features into a standardized canonical form on the fly, making the downstream processing more robust to variations such as scale, rotation, or translation. (D) Illustration of the pipeline for channel-specific spatial modulations: input $X$ induces different affine transformations across channels—e.g., channel $i$ undergoes scaling, channel $j$ experiences scaling and translation, and channel $k$ undergoes translation. (E) Input-dependent modulation of spatial receptive fields. The receptive field of the same neuron i is dynamically modulated based on different input stimuli X, Y, and Z. In these examples, the receptive field undergoes a scaling transformation for input X, a combination of scaling and translation for input Y, and translation for input Z.}
    \label{fig:theta12}
\end{figure*}

We introduce a novel readout that adaptively modulates both the encoder feature maps and their corresponding spatial weight distributions (i.e., receptive fields) on a per-voxel basis. Inspired by Spatial Transformer Networks (STN) \citep{jaderberg2015spatial}, this method spatially modulates the feature maps and spatial masks using affine transformations (e.g., rotation, scaling, and translation), allowing for dynamic and stimulus-dependent adjustments. The STN comprises two kinds of spatial modulations:

\textit{Spatial modulation of spatial masks (Receptive Fields).}
Unlike fixed spatial masks used in standard factorized or Gaussian readouts, our STN-based readout accommodates the dynamic nature of receptive fields (RFs). Biological evidence shows that RF sizes can expand or contract based on contrast \citep{sceniak1999contrast} and can also shift or reshape in response to contextual or attentional cues \citep{womelsdorf2006dynamic}. By allowing each voxel to learn its own affine transform, our method can capture such stimulus-dependent changes, moving beyond the static RF assumptions of conventional readouts (Figure \ref{fig:theta12} A, B, C).

\textit{Spatial modulation of feature maps.}
Beyond voxel-level RF modulation, STNs also enable channel-wise transformations of the encoder features. Each feature channel may encode distinct visual attributes (e.g., edges, textures, or shapes) and thus might require unique spatial modifications. In contrast to object classification tasks—where known invariances (e.g., rotation, reflection) can be applied through data augmentation—voxel responses exhibit unknown geometric invariances. Allowing the network to learn channel-specific transforms directly from fMRI data provides a powerful mechanism to discover these invariances, potentially leading to richer and more accurate neural response models (Figure \ref{fig:theta12} D, E). \\
\textbf{STN Architecture -} Our STN module has four key components: \textbf{1. Localization Network -} A pretrained ResNet-50 that processes the raw stimulus image and outputs a feature representation before adaptive average pooling, \textbf{2. Linear Deformation Networks -} Two linear networks produce affine transformation parameters. From the localization features, one generates \(\theta_1 \in \mathbb{R}^{C \times 6}\) for the \(C\) feature channels, while the other yields \(\theta_2 \in \mathbb{R}^{N \times 6}\) for the \(N\) voxels. Each row in \(\theta_1\) and \(\theta_2\) encodes a \(2\times3\) matrix (6 parameters) for a unique affine transform, \textbf{3. Parameterized Sampling Grid -} Constructs sampling grids based on \(\theta_1\) and \(\theta_2\), defining how \(\mathbf{E}\) (encoder feature map) and \(\mathbf{S}\) (spatial weight matrix) are warped and \textbf{4. Sampler -} Applies bilinear interpolation to generate the transformed feature map \(\mathbf{E}'\) and spatial weights \(\mathbf{S}'\).

We compute each voxel’s predicted response, \(\hat{\mathbf{Y}}_n\), using the Spatial-Feature Factorized Linear Readout (Eq.~\ref{eq:spatial_linear1}), but replace \(\mathbf{E}\) and \(\mathbf{S}\) with their STN-transformed versions:
\[
    \mathbf{E}' = \mathrm{AT}(\mathbf{E}, \theta_1), 
    \quad
    \mathbf{S}' = \mathrm{AT}(\mathbf{S}, \theta_2),
\]
where \(\mathrm{AT}(\mathbf{X}, \theta)\) applies a distinct \(2\times3\) affine matrix in \(\theta_m\) to each channel \(m\) in \(\mathbf{X} \in \mathbb{R}^{M \times W \times H}\). By jointly modulating receptive fields and feature channels, the STN readout captures the dynamic, context-dependent properties of neural responses and learns unknown geometric invariances directly from the data, offering a biologically motivated enhancement over fixed-mask readout methods.
Further analysis on this readout is expanded in Appendix Table \ref{tab:stn_sep_analysis}, Figure \ref{fig:affine_transform_alignment} and Section \nameref{subsec:stn_readouts}, where we examine how stimulus-dependent spatial shifts learned by the STN vary across the visual hierarchy. See Appendix Section~\nameref{subsec:stn_details}, Figures~\ref{fig:stn_summary} and Table~\ref{tab:complexity} for details on the individual affine transformations, computational complexity, and usability across different input stimuli.

\subsection{Training and Dataset}

In this study, we utilized stimuli-response pairs from four subjects (Subjects 1, 2, 5, and 7) from the Natural Scenes Dataset (More details in Appendix section \nameref{subsec:nsd}). The experimental setup involved presenting a total of 37,000 image stimuli from the MS COCO dataset \citep{lin2014microsoft} to these subjects. Out of these, 1,000 images were shown to all four subjects, and these shared images were designated as the test set for our analyses. The remaining 36,000 images were split into 35,000 for training and 1,000 for validation purposes. We trained separate models for each of the following brain regions: the high-level ventral, dorsal and lateral streams, V4, V3v, V3d, V2v, V2d, V1v, and V1d. This approach allowed us to tailor the models to the unique neural response patterns of each region, thereby providing a more precise understanding of how different parts of the visual cortex process information. Throughout the paper, the reported accuracy refers to the test-time performance, measured as the noise-normalized Pearson correlation between predicted and actual voxel responses (see Appendix section ~\nameref{subsec:nsd} for noise ceiling computation).

All response-optimized models were trained using an NVIDIA GeForce RTX 4090 and NVIDIA A40 GPU. We employed a batch size of 4 with gradient accumulation to achieve an effective batch size of 16, using a learning rate of 0.0001. Training was performed using an equal-weighted combination of Mean Squared Error (MSE) and correlation loss between predicted and target voxel responses, with early stopping applied after 20 epochs without improvement in validation accuracy, measured by Pearson correlation.

\section{Results}

\subsection{Performance comparison of readouts across vision and language models in the visual cortex}
\begin{figure}[h!]
\centering
\includegraphics[width=\linewidth, trim={0 10 0 0}, clip]{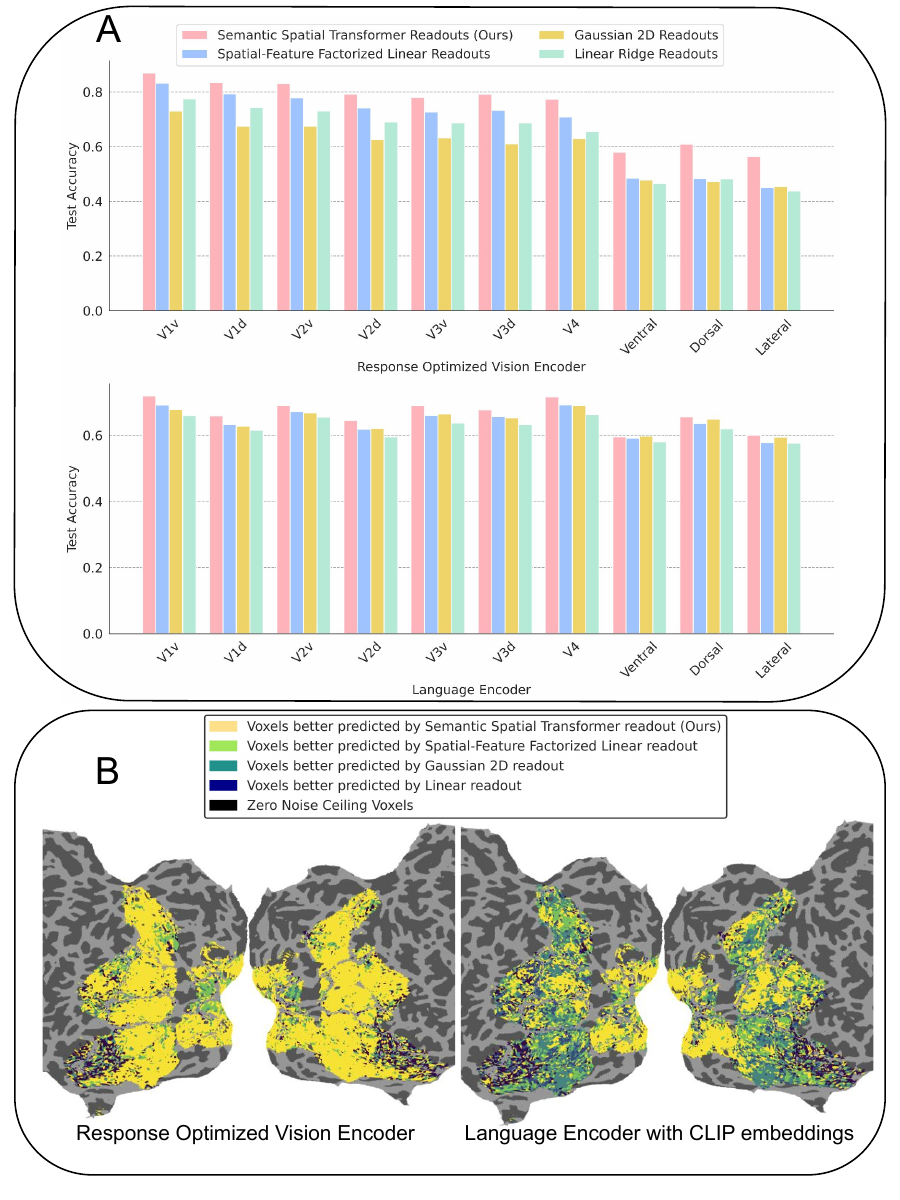}
\caption{Comparison of readout mechanisms - (A) Noise Normalized Test Accuracy (Pearson Correlation) on held out dataset for different brain regions calculated using Response-optimized vision and Dense Language (CLIP embedding) models using four different readouts , (B) Brain visualizations showing regions where each readout performs the best}
\label{fig:readout_comp}
\end{figure}

We first evaluated the performance of various readout mechanisms in predicting neural responses across different brain regions. Our results showed that the Semantic Spatial Transformer readouts consistently outperform Linear, 2D Gaussian, and Spatial-Feature Factorized Linear readouts across all regions of the visual cortex and for almost  all encoder models (see Figure~\ref{fig:readout_comp}). Its key advantage lies in the ability to flexibly adjust spatial masks and feature maps on a stimulus-by-stimulus basis, shifting receptive fields, resizing them, or rotating feature maps to align with a canonical form—transformations that better capture the actual variability in visual processing and boost predictive performance.
This trend of superior performance is especially evident in vision models (see Figure \ref{fig:readout_comp}-A) and holds for other task-optimized encoders processing visual input (details in Appendix Tables \ref{tab:Task_Optimized_model_comp} and \ref{tab:Response_Optimized_Vision_model_comp}). Figure \ref{fig:readout_comp}-B further illustrates the brain voxels where each readout performs best, underscoring the dominant performance of the Semantic Spatial Transformer readout for vision models across the visual hierarchy.

While the Semantic Spatial Transformer achieves the overall highest accuracy across all regions for all models (Appendix Tables \ref{tab:Task_Optimized_model_comp}, \ref{tab:Response_Optimized_Vision_model_comp}, \ref{tab:Response_Optimized_lANGUAGE_model_comp}), its improvement is less pronounced with language embedding inputs (Figure \ref{fig:readout_comp}-B). This disparity arises because the Semantic Spatial Transformer readout uses a pretrained ResNet50 encoder as the localization network to learn affine transformations that adjust both vision and language encoder feature spaces. Vision encoder features are generally larger per channel (e.g., 28×28) than language encoder features (e.g., 4×4). Consequently, the Semantic Spatial Transformer readout has a greater capacity to leverage the rich spatial information available in vision models. Larger spatial dimensions provide more granular information, allowing STNs to learn transformations that account for variations in position, scale, and orientation of features more accurately. Further analysis on this bias introduced by readouts can be found in Appendix section \nameref{subsec:stn_channel} and Table \ref{tab:stn_channel_analysis}.

Further, Spatial-Feature Factorized Linear Readouts outperform Linear Ridge Regression Readouts both in terms of memory efficiency and prediction performance, as shown in Figure \ref{fig:readout_comp}-A and Appendix Tables \ref{tab:Response_Optimized_lANGUAGE_model_comp}, \ref{tab:Response_Optimized_Vision_model_comp} and \ref{tab:Task_Optimized_model_comp}. This improvement is attributed to the readout's capability to effectively disentangle voxel response selectivity into spatial and feature dimensions. This approach aligns with established phenomena in neuroscience, where neurons exhibit selectivity not only for specific features but also for stimuli presented within their receptive field locations.

Gaussian 2D readouts are mostly outperformed by both Spatial-Feature Factorized Linear Readouts and Linear Ridge Regression Readouts in vision models, despite needing significantly fewer parameters. This performance gap can be attributed to the fact that Gaussian readouts were initially developed for grayscale stimuli in the mouse primary visual cortex~\citep{lurz2020generalization}, where they utilized the brain's retinotopic mapping and anatomical organization to accurately define receptive fields. In our study, however, we learn the parameters of the Gaussian readout solely from the responses to complex image inputs, deliberately excluding anatomical information to maintain a fair comparison with other methods. Furthermore, this modeling approach may be less effective for the human visual system, where the assumption of a Gaussian-like structure may not hold true for the spatial receptive fields of all voxels, which may exhibit greater complexity.

Interestingly, the performance gap between Gaussian readouts and other readouts narrows in language models, where Gaussian readouts slightly outperform linear readouts across all regions and exceed Spatial-Feature Factorized Linear Readouts in higher regions. This may be due to the smaller feature space in language models compared to vision models (e.g., 4×4 vs. 28×28), which simplifies receptive field localization. 

Since the Semantic Spatial Transformer readout consistently outperformed others, we focus on it when analyzing the encoders in detail in the following sections.

\begin{figure}[h!]
\centering
\includegraphics[width=\linewidth, trim={0 140 0 0}, clip]{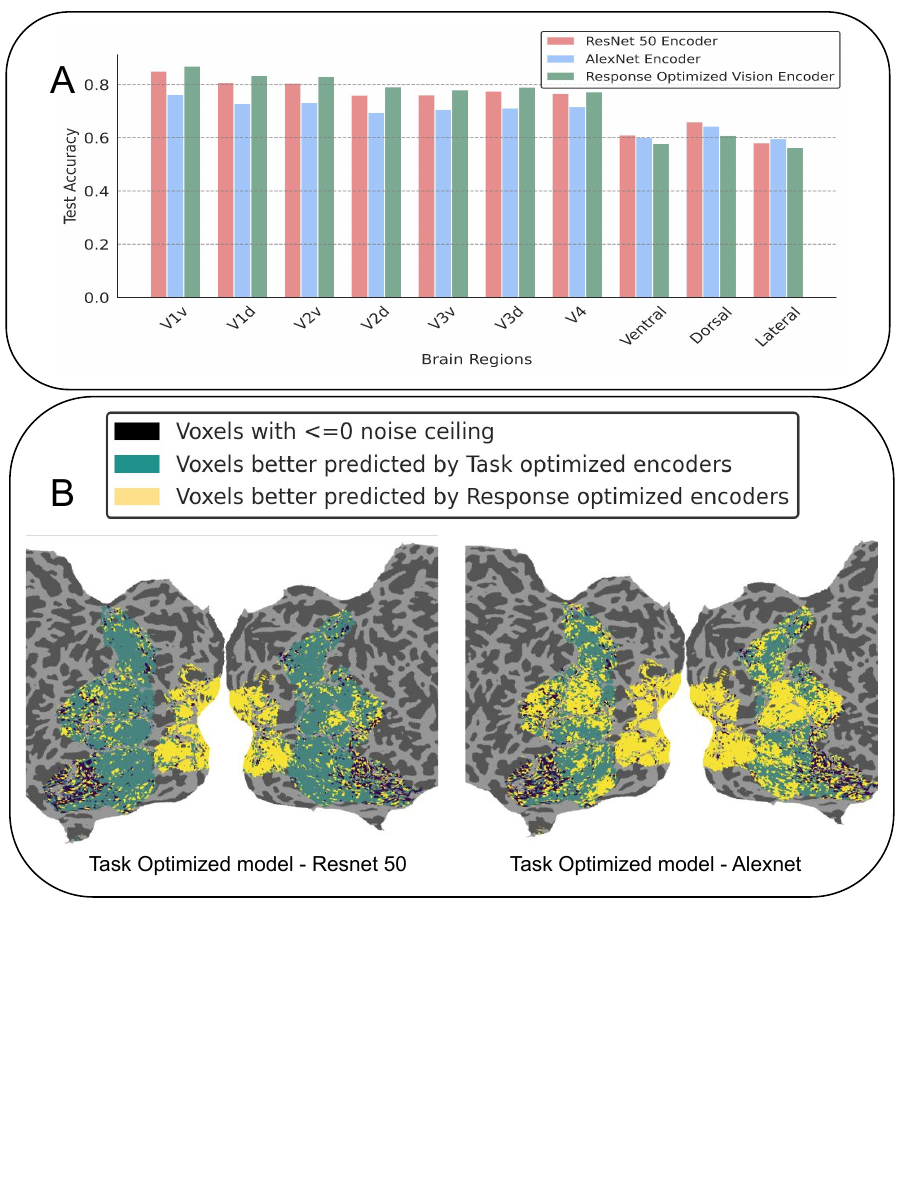}

\caption{Comparison of Task-optimized versus Response-optimized vision models - (A) Test Accuracy (Normalized Pearson Correlation) on held out dataset using Task-optimized model encoders and Response-optimized model encoders with Semantic Spatial Transformer readout, (B) Brain visualization showing voxels better predicted by each model}
\label{fig:rpo_task_all}
\end{figure}

\begin{table*}[!ht]
\begin{center} 
\caption{Performance (Test Accuracies as Noise-Normalized Pearson Correlation) of Task-Optimized Vision models (ResNet-50, TV; best from \ref{tab:Task_Optimized_model_comp}), Response-Optimized Vision models (RV), and Language Models with CLIP embeddings—Single Caption (SL) and Dense Caption (DL). All use the Semantic Spatial Transformer readout, except SL which uses Ridge Linear readout.} 
\label{tab:Results_summary} 
\vskip 0.12in
\begin{tabular}{|c|c|c|c|c|c|c|c|c|c|c|}
\hline
\textbf{Model Details} & \textbf{V1v} & \textbf{V1d} & \textbf{V2v} & \textbf{V2d} & \textbf{V3v} & \textbf{V3d} & \textbf{V4} & \textbf{Ventral} & \textbf{Dorsal} & \textbf{Lateral} \\
\hline
TV & 0.8507 & 0.8083 & 0.8057 & 0.7603 & 0.7612 & 0.7763 & 0.7674 & \textbf{0.6105} & \textbf{0.6606} & 0.5823 \\
RV & \textbf{0.8698} & \textbf{0.8340} & \textbf{0.8302} & \textbf{0.7919} & \textbf{0.7808} & \textbf{0.7913} & \textbf{0.7729} & 0.5796 & 0.6089 & 0.5638 \\
SL & 0.3974 & 0.3779 & 0.3809 & 0.3702 & 0.4093 & 0.4119 & 0.4882 & 0.5661 & 0.6243 & 0.5920 \\
DL & 0.7196 & 0.6590 & 0.6903 & 0.6457 & 0.6897 & 0.6774 & 0.7167 & 0.5953 & 0.6562 & \textbf{0.6001} \\
\hline
\end{tabular} 
\end{center} 
\end{table*}

\begin{figure*}[h!]
\centering
\includegraphics[width=\linewidth, trim={0 120 110 0}, clip]{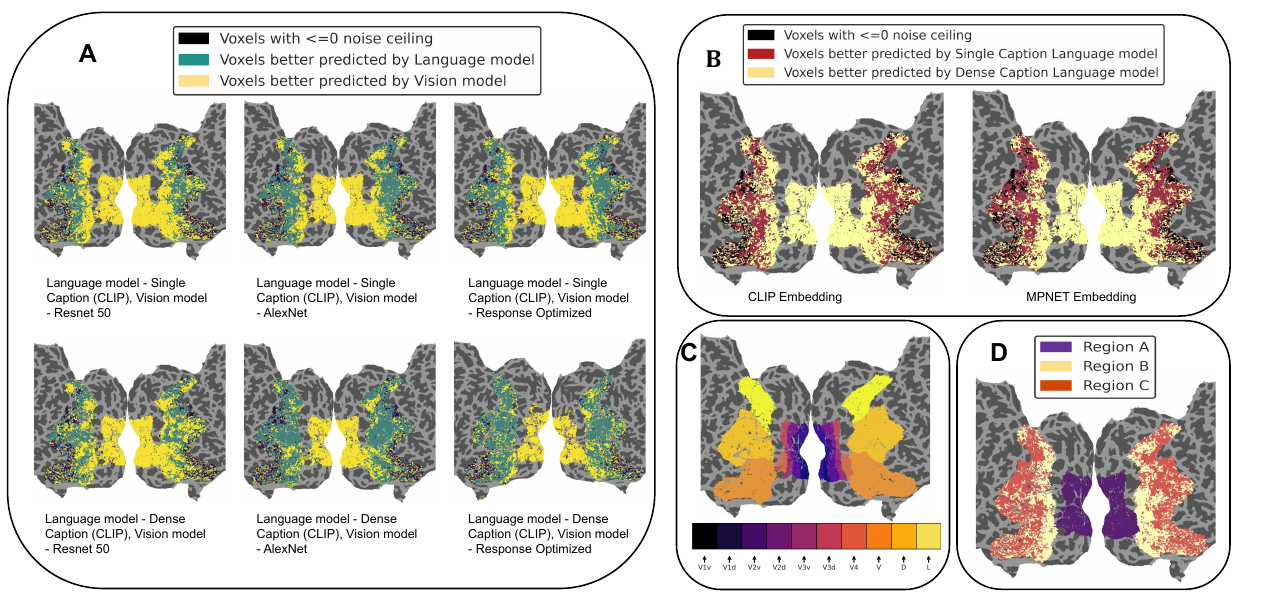}
\caption{Comparison of vision and language models using Semantic Spatial Transformer readouts - Brain visualizations showing: (A) voxels better predicted by vision and language models, (B) voxels better predicted by single and dense caption language models, (C) the ten regions of the human visual cortex analysed in this study (V, D and L refer to Ventral, Dorsal and Lateral streams respectively), (D) highlighting three distinct regions, each demonstrating varying sensitivities to largely perceptual characteristics of the input, localized visual semantics aligned with linguistic descriptions, and global
semantic interpretations of the input, also aligned with language}
\label{fig:vision_lang}
\end{figure*}

\subsection{Task-optimized vs Response-optimized models}

To ensure a fair comparison, we trained models using different sets of layers for each task-optimized model (Appendix Table \ref{tab:Task_Optimized_model_comp} containing additional baselines ConvNext-Base \cite{liu2022convnet} and MOCO-V2 \cite{he2020momentum}), and used only the best-performing ResNet50 layers for comparison, as presented in Table \ref{tab:Results_summary}. In the early regions of the visual cortex (V1, V2, V3, and V4), response-optimized vision models consistently outperform task-optimized models by 2-12$\%$ (Figure \ref{fig:rpo_task_all} and Table \ref{tab:Results_summary}), with a particularly notable margin over simpler architectures like AlexNet (Appendix Table \ref{tab:Task_Optimized_model_comp}).  This suggests that features necessary for modeling early and mid-level visual areas are not fully captured by current task-optimized models, and explicit alignment with neural responsefs is crucial for higher prediction accuracy. This may be because task-optimized models, primarily trained on object-centric tasks, don't account for the broader range of visual functions performed by the brain. Incorporating more ethologically relevant tasks into the optimization framework might be necessary for better modeling of early to mid-level visual processing. In the higher regions of the visual cortex (high-level ventral, dorsal, and lateral streams), task-optimized models show a slight performance advantage of around 5$\%$ over response-optimized models. This could be because these regions process more complex visual information, and task-optimized models, trained on larger object-centric datasets like ImageNet ($\geq$1.2 million images), better capture these functions. However, the small difference indicates that response-optimized models, despite being trained on only a fraction (~3\%) of the data, still capture significant aspects of high-level visual processing.

\subsection{Brain regions sensitive to vision vs language models}

Recent research shows that pure language models, like MPNET, can predict image-evoked brain activity in the high-level visual cortex using only image captions~\citep{doerig2024Visual}. This raises intriguing questions about the alignment between the human visual cortex and language. To explore this relationship further, we compare these language  with vision-only models.

When we assess language models that receive only image captions—without the images themselves—against response-optimized vision models, we find that the lower regions of the visual cortex are better modeled by vision-based approaches. In contrast, higher regions are more effectively captured by language models (see Figure \ref{fig:vision_lang}-A, column 3 and Table \ref{tab:Results_summary}). This pattern also holds when comparing language models to task-optimized vision models, although the distinction is less pronounced (first two columns of Figure \ref{fig:vision_lang}-A).

Next, we differentiate between single-caption and dense-caption models. Single-caption models convey only the overall semantic content of an image, whereas dense-caption models capture both spatial and semantic details. Consequently, the lower regions of the visual cortex, which are sensitive to fine-grained visual information, are better modeled by dense-caption models, as illustrated in Figure \ref{fig:vision_lang}-B.

As we move from the lower to the higher regions of the visual cortex, there is a notable shift in sensitivity from localized semantics to global semantics across all ventral, dorsal, and lateral streams. Figure \ref{fig:vision_lang}-B demonstrates that single-caption models dominate in the mid-to-higher regions of these streams, emphasizing the sensitivity of these areas to the overall meaning or interpretation of an entire image or scene. This trend is further corroborated in Figure \ref{fig:vision_lang}-A, which compares vision models with both single-caption and dense-caption language models. Here, response-optimized vision models outperform single-caption models in the lower regions of the ventral, dorsal, and lateral streams, but do not maintain this advantage in the mid-to-higher regions.

Thus, we can identify three distinct regions in the visual cortex that are sensitive to different stimulus types (Figure \ref{fig:vision_lang}-D): (1) lower visual regions (V1, V2, V3, and V4) are most sensitive to perceptual features that are not fully captured by linguistic descriptions - region A; (2) mid-level regions of the dorsal, ventral, and lateral streams are most sensitive to localized semantics (i.e. detailed, specific information about particular parts or regions of an image) - region B; and (3) higher regions of the dorsal, ventral, and lateral streams are sensitive exclusively to global semantic information - region C. Vision models outperform both single and dense caption language models in region A (Figure \ref{fig:vision_lang}-A and Table \ref{tab:Results_summary}), thus proving its sensitivity to largely perceptual features. Dense Caption language models outperform single caption language models (Figure \ref{fig:vision_lang}-B) and response-optimized vision models (Figure \ref{fig:vision_lang}-A) in region B, thus proving it is most sensitive to nuanced, localized semantic details. Vision models also outperform single caption models in region B (Figure \ref{fig:vision_lang}-A), thus proving it is more sensitive to detailed visual information. Lastly, single caption language models outperform both dense caption models (Figure \ref{fig:vision_lang}-B) and vision models (Figure \ref{fig:vision_lang}-A) in region C, thus confirming its sensitivity to global semantics. Although this comparison was done mainly using Semantic Spatial Transformer readout, the trends hold true for other readouts, although to a much lesser extent (Appendix Figures \ref{fig:vision_lang_spatial_linear}, \ref{fig:vision_lang_gaussian2D}, \ref{fig:vision_lang_ridge_linear}).

\section{Discussion}

In this study, we leveraged the NSD Dataset to evaluate various neural network models in predicting neural responses across different brain regions. Our analysis focused on three key comparisons: task vs. response optimized models, vision models vs. language models, and different readout methods for mapping model activations to brain signals.

First, we compared task-optimized models pre-trained on visual tasks (thus biased toward those tasks), with response-optimized models trained directly from brain response data. Response-optimized models significantly outperform task-optimized models in early visual regions. This suggests that brain-like processing in early-to-mid visual areas does not fully emerge in task-optimized models, and explicit alignment with neural data enhances prediction accuracy. However, in higher visual regions, both model types perform comparably, with task-optimized models showing a slight edge.

Next, we compared vision models with language models (both single-caption and dense-caption). Vision models outperformed language models in early visual regions, which are more attuned to perceptual features not captured by linguistic descriptions. In mid-level visual regions, sensitivity shifts toward semantic information, with dense-caption models excelling due to their ability to represent localized semantics. In higher visual regions, single-caption models perform better, indicating the importance of global scene understanding.

Finally, we evaluated different readout mechanisms for mapping activations to brain responses. Factorized readouts significantly outperformed standard linear models, and incorporating a Semantic Spatial Transformer further improved performance, particularly in vision models.

Our work has several limitations. First, we focused on task-optimized models trained for object categorization. A comprehensive comparison of models trained on other visual objectives and data sets is outside the scope of this study. However, prior research suggests that variations in architecture, objective, and data diet do not drastically impact response prediction accuracy~\citep{conwell2022can}, so we do not expect our conclusions to change significantly with additional models. While we found that language models become more accurate in predicting responses in high-level visual regions, we did not explore what specifically drives this performance \cite{shoham2024organization}, \cite{conwell2023unreasonable}, \cite{huh2024platonic}. It is still uncertain whether object category information (e.g., nouns) or other elements such as actions, spatial relationships, or contextual details play a more significant role. Finally, while the Semantic Spatial Transformer led to better predictions, future work should investigate how spatial and feature weights are modulated by different inputs. We also only tested affine transformations; more constrained or nonlinear deformations may offer further improvements.

\newpage

\bibliographystyle{ccn_style}

\bibliography{ccn_style}

\begin{thebibliography}{}

\bibitem [\protect \citeauthoryear {%
Abdelhack%
\ \BBA {} Kamitani%
}{%
Abdelhack%
\ \BBA {} Kamitani%
}{%
{\protect \APACyear {2018}}%
}]{%
abdelhack2018sharpening}
\APACinsertmetastar {%
abdelhack2018sharpening}%
\begin{APACrefauthors}%
Abdelhack, M.%
\BCBT {}\ \BBA {} Kamitani, Y.%
\end{APACrefauthors}%
\unskip\
\newblock
\APACrefYearMonthDay{2018}{}{}.
\newblock
{\BBOQ}\APACrefatitle {Sharpening of hierarchical visual feature representations of blurred images} {Sharpening of hierarchical visual feature representations of blurred images}.{\BBCQ}
\newblock
\APACjournalVolNumPages{eneuro}{5}{3}{}.
\PrintBackRefs{\CurrentBib}

\bibitem [\protect \citeauthoryear {%
Albrecht%
\ \BBA {} Hamilton%
}{%
Albrecht%
\ \BBA {} Hamilton%
}{%
{\protect \APACyear {1982}}%
}]{%
albrecht1982striate}
\APACinsertmetastar {%
albrecht1982striate}%
\begin{APACrefauthors}%
Albrecht, D\BPBI G.%
\BCBT {}\ \BBA {} Hamilton, D\BPBI B.%
\end{APACrefauthors}%
\unskip\
\newblock
\APACrefYearMonthDay{1982}{}{}.
\newblock
{\BBOQ}\APACrefatitle {Striate cortex of monkey and cat: contrast response function.} {Striate cortex of monkey and cat: contrast response function.}{\BBCQ}
\newblock
\APACjournalVolNumPages{Journal of neurophysiology}{48}{1}{217--237}.
\PrintBackRefs{\CurrentBib}

\bibitem [\protect \citeauthoryear {%
Allen%
\ \protect \BOthers {.}}{%
Allen%
\ \protect \BOthers {.}}{%
{\protect \APACyear {2022}}%
}]{%
allen2022massive}
\APACinsertmetastar {%
allen2022massive}%
\begin{APACrefauthors}%
Allen, E\BPBI J.%
, St-Yves, G.%
, Wu, Y.%
, Breedlove, J\BPBI L.%
, Prince, J\BPBI S.%
, Dowdle, L\BPBI T.%
\BDBL {}others%
\end{APACrefauthors}%
\unskip\
\newblock
\APACrefYearMonthDay{2022}{}{}.
\newblock
{\BBOQ}\APACrefatitle {A massive 7T fMRI dataset to bridge cognitive neuroscience and artificial intelligence} {A massive 7t fmri dataset to bridge cognitive neuroscience and artificial intelligence}.{\BBCQ}
\newblock
\APACjournalVolNumPages{Nature neuroscience}{25}{1}{116--126}.
\PrintBackRefs{\CurrentBib}

\bibitem [\protect \citeauthoryear {%
Bashivan%
, Kar%
\BCBL {}\ \BBA {} DiCarlo%
}{%
Bashivan%
\ \protect \BOthers {.}}{%
{\protect \APACyear {2019}}%
}]{%
bashivan2019neural}
\APACinsertmetastar {%
bashivan2019neural}%
\begin{APACrefauthors}%
Bashivan, P.%
, Kar, K.%
\BCBL {}\ \BBA {} DiCarlo, J\BPBI J.%
\end{APACrefauthors}%
\unskip\
\newblock
\APACrefYearMonthDay{2019}{}{}.
\newblock
{\BBOQ}\APACrefatitle {Neural population control via deep image synthesis} {Neural population control via deep image synthesis}.{\BBCQ}
\newblock
\APACjournalVolNumPages{Science}{364}{6439}{eaav9436}.
\PrintBackRefs{\CurrentBib}

\bibitem [\protect \citeauthoryear {%
Brincat%
\ \BBA {} Connor%
}{%
Brincat%
\ \BBA {} Connor%
}{%
{\protect \APACyear {2004}}%
}]{%
brincat2004underlying}
\APACinsertmetastar {%
brincat2004underlying}%
\begin{APACrefauthors}%
Brincat, S\BPBI L.%
\BCBT {}\ \BBA {} Connor, C\BPBI E.%
\end{APACrefauthors}%
\unskip\
\newblock
\APACrefYearMonthDay{2004}{}{}.
\newblock
{\BBOQ}\APACrefatitle {Underlying principles of visual shape selectivity in posterior inferotemporal cortex} {Underlying principles of visual shape selectivity in posterior inferotemporal cortex}.{\BBCQ}
\newblock
\APACjournalVolNumPages{Nature neuroscience}{7}{8}{880--886}.
\PrintBackRefs{\CurrentBib}

\bibitem [\protect \citeauthoryear {%
Brown%
\ \protect \BOthers {.}}{%
Brown%
\ \protect \BOthers {.}}{%
{\protect \APACyear {2020}}%
}]{%
brown2020language}
\APACinsertmetastar {%
brown2020language}%
\begin{APACrefauthors}%
Brown, T.%
, Mann, B.%
, Ryder, N.%
, Subbiah, M.%
, Kaplan, J\BPBI D.%
, Dhariwal, P.%
\BDBL {}others%
\end{APACrefauthors}%
\unskip\
\newblock
\APACrefYearMonthDay{2020}{}{}.
\newblock
{\BBOQ}\APACrefatitle {Language models are few-shot learners} {Language models are few-shot learners}.{\BBCQ}
\newblock
\APACjournalVolNumPages{Advances in neural information processing systems}{33}{}{1877--1901}.
\PrintBackRefs{\CurrentBib}

\bibitem [\protect \citeauthoryear {%
Cichy%
, Khosla%
, Pantazis%
, Torralba%
\BCBL {}\ \BBA {} Oliva%
}{%
Cichy%
\ \protect \BOthers {.}}{%
{\protect \APACyear {2016}}%
}]{%
cichy2016comparison}
\APACinsertmetastar {%
cichy2016comparison}%
\begin{APACrefauthors}%
Cichy, R\BPBI M.%
, Khosla, A.%
, Pantazis, D.%
, Torralba, A.%
\BCBL {}\ \BBA {} Oliva, A.%
\end{APACrefauthors}%
\unskip\
\newblock
\APACrefYearMonthDay{2016}{}{}.
\newblock
{\BBOQ}\APACrefatitle {Comparison of deep neural networks to spatio-temporal cortical dynamics of human visual object recognition reveals hierarchical correspondence} {Comparison of deep neural networks to spatio-temporal cortical dynamics of human visual object recognition reveals hierarchical correspondence}.{\BBCQ}
\newblock
\APACjournalVolNumPages{Scientific reports}{6}{1}{27755}.
\PrintBackRefs{\CurrentBib}

\bibitem [\protect \citeauthoryear {%
Conwell%
, Prince%
, Alvarez%
\BCBL {}\ \BBA {} Konkle%
}{%
Conwell%
\ \protect \BOthers {.}}{%
{\protect \APACyear {2023}}%
}]{%
conwell2023unreasonable}
\APACinsertmetastar {%
conwell2023unreasonable}%
\begin{APACrefauthors}%
Conwell, C.%
, Prince, J.%
, Alvarez, G.%
\BCBL {}\ \BBA {} Konkle, T.%
\end{APACrefauthors}%
\unskip\
\newblock
\APACrefYearMonthDay{2023}{}{}.
\newblock
{\BBOQ}\APACrefatitle {The unreasonable effectiveness of word models in predicting high-level visual cortex responses to natural images} {The unreasonable effectiveness of word models in predicting high-level visual cortex responses to natural images}.{\BBCQ}
\newblock
\BIn{} \APACrefbtitle {Conference on computational cognitive neuroscience.} {Conference on computational cognitive neuroscience.}
\PrintBackRefs{\CurrentBib}

\bibitem [\protect \citeauthoryear {%
Conwell%
, Prince%
, Alvarez%
\BCBL {}\ \BBA {} Konkle%
}{%
Conwell%
, Prince%
, Alvarez%
\BCBL {}\ \BBA {} Konkle%
}{%
{\protect \APACyear {2022}}%
}]{%
conwell2022large}
\APACinsertmetastar {%
conwell2022large}%
\begin{APACrefauthors}%
Conwell, C.%
, Prince, J\BPBI S.%
, Alvarez, G\BPBI A.%
\BCBL {}\ \BBA {} Konkle, T.%
\end{APACrefauthors}%
\unskip\
\newblock
\APACrefYearMonthDay{2022}{}{}.
\newblock
{\BBOQ}\APACrefatitle {Large-scale benchmarking of diverse artificial vision models in prediction of 7T human neuroimaging data} {Large-scale benchmarking of diverse artificial vision models in prediction of 7t human neuroimaging data}.{\BBCQ}
\newblock
\APACjournalVolNumPages{BioRxiv}{}{}{}.
\PrintBackRefs{\CurrentBib}

\bibitem [\protect \citeauthoryear {%
Conwell%
, Prince%
, Kay%
, Alvarez%
\BCBL {}\ \BBA {} Konkle%
}{%
Conwell%
, Prince%
, Kay%
\BCBL {}\ \protect \BOthers {.}}{%
{\protect \APACyear {2022}}%
}]{%
conwell2022can}
\APACinsertmetastar {%
conwell2022can}%
\begin{APACrefauthors}%
Conwell, C.%
, Prince, J\BPBI S.%
, Kay, K\BPBI N.%
, Alvarez, G\BPBI A.%
\BCBL {}\ \BBA {} Konkle, T.%
\end{APACrefauthors}%
\unskip\
\newblock
\APACrefYearMonthDay{2022}{}{}.
\newblock
{\BBOQ}\APACrefatitle {What can 1.8 billion regressions tell us about the pressures shaping high-level visual representation in brains and machines?} {What can 1.8 billion regressions tell us about the pressures shaping high-level visual representation in brains and machines?}{\BBCQ}
\newblock
\APACjournalVolNumPages{BioRxiv}{}{}{2022--03}.
\PrintBackRefs{\CurrentBib}

\bibitem [\protect \citeauthoryear {%
Conwell%
, Prince%
, Kay%
, Alvarez%
\BCBL {}\ \BBA {} Konkle%
}{%
Conwell%
\ \protect \BOthers {.}}{%
{\protect \APACyear {2024}}%
}]{%
conwell2024large}
\APACinsertmetastar {%
conwell2024large}%
\begin{APACrefauthors}%
Conwell, C.%
, Prince, J\BPBI S.%
, Kay, K\BPBI N.%
, Alvarez, G\BPBI A.%
\BCBL {}\ \BBA {} Konkle, T.%
\end{APACrefauthors}%
\unskip\
\newblock
\APACrefYearMonthDay{2024}{}{}.
\newblock
{\BBOQ}\APACrefatitle {A large-scale examination of inductive biases shaping high-level visual representation in brains and machines} {A large-scale examination of inductive biases shaping high-level visual representation in brains and machines}.{\BBCQ}
\newblock
\APACjournalVolNumPages{Nature communications}{15}{1}{9383}.
\PrintBackRefs{\CurrentBib}

\bibitem [\protect \citeauthoryear {%
Dapello%
\ \protect \BOthers {.}}{%
Dapello%
\ \protect \BOthers {.}}{%
{\protect \APACyear {2022}}%
}]{%
dapello2022aligning}
\APACinsertmetastar {%
dapello2022aligning}%
\begin{APACrefauthors}%
Dapello, J.%
, Kar, K.%
, Schrimpf, M.%
, Geary, R.%
, Ferguson, M.%
, Cox, D\BPBI D.%
\BCBL {}\ \BBA {} DiCarlo, J\BPBI J.%
\end{APACrefauthors}%
\unskip\
\newblock
\APACrefYearMonthDay{2022}{}{}.
\newblock
{\BBOQ}\APACrefatitle {Aligning model and macaque inferior temporal cortex representations improves model-to-human behavioral alignment and adversarial robustness} {Aligning model and macaque inferior temporal cortex representations improves model-to-human behavioral alignment and adversarial robustness}.{\BBCQ}
\newblock
\APACjournalVolNumPages{bioRxiv}{}{}{2022--07}.
\PrintBackRefs{\CurrentBib}

\bibitem [\protect \citeauthoryear {%
Deng%
\ \protect \BOthers {.}}{%
Deng%
\ \protect \BOthers {.}}{%
{\protect \APACyear {2009}}%
}]{%
deng2009imagenet}
\APACinsertmetastar {%
deng2009imagenet}%
\begin{APACrefauthors}%
Deng, J.%
, Dong, W.%
, Socher, R.%
, Li, L\BHBI J.%
, Li, K.%
\BCBL {}\ \BBA {} Fei-Fei, L.%
\end{APACrefauthors}%
\unskip\
\newblock
\APACrefYearMonthDay{2009}{}{}.
\newblock
{\BBOQ}\APACrefatitle {Imagenet: A large-scale hierarchical image database} {Imagenet: A large-scale hierarchical image database}.{\BBCQ}
\newblock
\BIn{} \APACrefbtitle {2009 IEEE conference on computer vision and pattern recognition} {2009 ieee conference on computer vision and pattern recognition}\ (\BPGS\ 248--255).
\PrintBackRefs{\CurrentBib}

\bibitem [\protect \citeauthoryear {%
Desimone%
, Albright%
, Gross%
\BCBL {}\ \BBA {} Bruce%
}{%
Desimone%
\ \protect \BOthers {.}}{%
{\protect \APACyear {1984}}%
}]{%
desimone1984stimulus}
\APACinsertmetastar {%
desimone1984stimulus}%
\begin{APACrefauthors}%
Desimone, R.%
, Albright, T\BPBI D.%
, Gross, C\BPBI G.%
\BCBL {}\ \BBA {} Bruce, C.%
\end{APACrefauthors}%
\unskip\
\newblock
\APACrefYearMonthDay{1984}{}{}.
\newblock
{\BBOQ}\APACrefatitle {Stimulus-selective properties of inferior temporal neurons in the macaque} {Stimulus-selective properties of inferior temporal neurons in the macaque}.{\BBCQ}
\newblock
\APACjournalVolNumPages{Journal of Neuroscience}{4}{8}{2051--2062}.
\PrintBackRefs{\CurrentBib}

\bibitem [\protect \citeauthoryear {%
Doerig%
\ \protect \BOthers {.}}{%
Doerig%
\ \protect \BOthers {.}}{%
{\protect \APACyear {2024}}%
}]{%
doerig2024Visual}
\APACinsertmetastar {%
doerig2024Visual}%
\begin{APACrefauthors}%
Doerig, A.%
, Kietzmann, T\BPBI C.%
, Allen, E.%
, Wu, Y.%
, Naselaris, T.%
, Kay, K.%
\BCBL {}\ \BBA {} Charest, I.%
\end{APACrefauthors}%
\unskip\
\newblock
\APACrefYearMonthDay{2024}{}{}.
\newblock
{\BBOQ}\APACrefatitle {Visual representations in the human brain are aligned with large language models} {Visual representations in the human brain are aligned with large language models}.{\BBCQ}
\newblock
\APACjournalVolNumPages{arXiv preprint arXiv:2209.11737}{}{}{}.
\PrintBackRefs{\CurrentBib}

\bibitem [\protect \citeauthoryear {%
Eickenberg%
, Gramfort%
, Varoquaux%
\BCBL {}\ \BBA {} Thirion%
}{%
Eickenberg%
\ \protect \BOthers {.}}{%
{\protect \APACyear {2017}}%
}]{%
eickenberg2017seeing}
\APACinsertmetastar {%
eickenberg2017seeing}%
\begin{APACrefauthors}%
Eickenberg, M.%
, Gramfort, A.%
, Varoquaux, G.%
\BCBL {}\ \BBA {} Thirion, B.%
\end{APACrefauthors}%
\unskip\
\newblock
\APACrefYearMonthDay{2017}{}{}.
\newblock
{\BBOQ}\APACrefatitle {Seeing it all: Convolutional network layers map the function of the human visual system} {Seeing it all: Convolutional network layers map the function of the human visual system}.{\BBCQ}
\newblock
\APACjournalVolNumPages{NeuroImage}{152}{}{184--194}.
\PrintBackRefs{\CurrentBib}

\bibitem [\protect \citeauthoryear {%
Federer%
, Xu%
, Fyshe%
\BCBL {}\ \BBA {} Zylberberg%
}{%
Federer%
\ \protect \BOthers {.}}{%
{\protect \APACyear {2020}}%
}]{%
federer2020improved}
\APACinsertmetastar {%
federer2020improved}%
\begin{APACrefauthors}%
Federer, C.%
, Xu, H.%
, Fyshe, A.%
\BCBL {}\ \BBA {} Zylberberg, J.%
\end{APACrefauthors}%
\unskip\
\newblock
\APACrefYearMonthDay{2020}{}{}.
\newblock
{\BBOQ}\APACrefatitle {Improved object recognition using neural networks trained to mimic the brain’s statistical properties} {Improved object recognition using neural networks trained to mimic the brain’s statistical properties}.{\BBCQ}
\newblock
\APACjournalVolNumPages{Neural Networks}{131}{}{103--114}.
\PrintBackRefs{\CurrentBib}

\bibitem [\protect \citeauthoryear {%
Gallant%
, Braun%
\BCBL {}\ \BBA {} Van~Essen%
}{%
Gallant%
\ \protect \BOthers {.}}{%
{\protect \APACyear {1993}}%
}]{%
gallant1993selectivity}
\APACinsertmetastar {%
gallant1993selectivity}%
\begin{APACrefauthors}%
Gallant, J\BPBI L.%
, Braun, J.%
\BCBL {}\ \BBA {} Van~Essen, D\BPBI C.%
\end{APACrefauthors}%
\unskip\
\newblock
\APACrefYearMonthDay{1993}{}{}.
\newblock
{\BBOQ}\APACrefatitle {Selectivity for polar, hyperbolic, and Cartesian gratings in macaque visual cortex} {Selectivity for polar, hyperbolic, and cartesian gratings in macaque visual cortex}.{\BBCQ}
\newblock
\APACjournalVolNumPages{Science}{259}{5091}{100--103}.
\PrintBackRefs{\CurrentBib}

\bibitem [\protect \citeauthoryear {%
G{\"u}{\c{c}}l{\"u}%
\ \BBA {} Van~Gerven%
}{%
G{\"u}{\c{c}}l{\"u}%
\ \BBA {} Van~Gerven%
}{%
{\protect \APACyear {2015}}%
}]{%
gucclu2015deep}
\APACinsertmetastar {%
gucclu2015deep}%
\begin{APACrefauthors}%
G{\"u}{\c{c}}l{\"u}, U.%
\BCBT {}\ \BBA {} Van~Gerven, M\BPBI A.%
\end{APACrefauthors}%
\unskip\
\newblock
\APACrefYearMonthDay{2015}{}{}.
\newblock
{\BBOQ}\APACrefatitle {Deep neural networks reveal a gradient in the complexity of neural representations across the ventral stream} {Deep neural networks reveal a gradient in the complexity of neural representations across the ventral stream}.{\BBCQ}
\newblock
\APACjournalVolNumPages{Journal of Neuroscience}{35}{27}{10005--10014}.
\PrintBackRefs{\CurrentBib}

\bibitem [\protect \citeauthoryear {%
He%
, Fan%
, Wu%
, Xie%
\BCBL {}\ \BBA {} Girshick%
}{%
He%
\ \protect \BOthers {.}}{%
{\protect \APACyear {2020}}%
}]{%
he2020momentum}
\APACinsertmetastar {%
he2020momentum}%
\begin{APACrefauthors}%
He, K.%
, Fan, H.%
, Wu, Y.%
, Xie, S.%
\BCBL {}\ \BBA {} Girshick, R.%
\end{APACrefauthors}%
\unskip\
\newblock
\APACrefYearMonthDay{2020}{}{}.
\newblock
{\BBOQ}\APACrefatitle {Momentum contrast for unsupervised visual representation learning} {Momentum contrast for unsupervised visual representation learning}.{\BBCQ}
\newblock
\BIn{} \APACrefbtitle {Proceedings of the IEEE/CVF conference on computer vision and pattern recognition} {Proceedings of the ieee/cvf conference on computer vision and pattern recognition}\ (\BPGS\ 9729--9738).
\PrintBackRefs{\CurrentBib}

\bibitem [\protect \citeauthoryear {%
He%
, Zhang%
, Ren%
\BCBL {}\ \BBA {} Sun%
}{%
He%
\ \protect \BOthers {.}}{%
{\protect \APACyear {2016}}%
}]{%
he2016deep}
\APACinsertmetastar {%
he2016deep}%
\begin{APACrefauthors}%
He, K.%
, Zhang, X.%
, Ren, S.%
\BCBL {}\ \BBA {} Sun, J.%
\end{APACrefauthors}%
\unskip\
\newblock
\APACrefYearMonthDay{2016}{}{}.
\newblock
{\BBOQ}\APACrefatitle {Deep residual learning for image recognition} {Deep residual learning for image recognition}.{\BBCQ}
\newblock
\BIn{} \APACrefbtitle {Proceedings of the IEEE conference on computer vision and pattern recognition} {Proceedings of the ieee conference on computer vision and pattern recognition}\ (\BPGS\ 770--778).
\PrintBackRefs{\CurrentBib}

\bibitem [\protect \citeauthoryear {%
Horikawa%
\ \BBA {} Kamitani%
}{%
Horikawa%
\ \BBA {} Kamitani%
}{%
{\protect \APACyear {2017}}%
}]{%
horikawa2017generic}
\APACinsertmetastar {%
horikawa2017generic}%
\begin{APACrefauthors}%
Horikawa, T.%
\BCBT {}\ \BBA {} Kamitani, Y.%
\end{APACrefauthors}%
\unskip\
\newblock
\APACrefYearMonthDay{2017}{}{}.
\newblock
{\BBOQ}\APACrefatitle {Generic decoding of seen and imagined objects using hierarchical visual features} {Generic decoding of seen and imagined objects using hierarchical visual features}.{\BBCQ}
\newblock
\APACjournalVolNumPages{Nature communications}{8}{1}{15037}.
\PrintBackRefs{\CurrentBib}

\bibitem [\protect \citeauthoryear {%
Hubel%
\ \BBA {} Wiesel%
}{%
Hubel%
\ \BBA {} Wiesel%
}{%
{\protect \APACyear {1962}}%
}]{%
hubel1962receptive}
\APACinsertmetastar {%
hubel1962receptive}%
\begin{APACrefauthors}%
Hubel, D\BPBI H.%
\BCBT {}\ \BBA {} Wiesel, T\BPBI N.%
\end{APACrefauthors}%
\unskip\
\newblock
\APACrefYearMonthDay{1962}{}{}.
\newblock
{\BBOQ}\APACrefatitle {Receptive fields, binocular interaction and functional architecture in the cat's visual cortex} {Receptive fields, binocular interaction and functional architecture in the cat's visual cortex}.{\BBCQ}
\newblock
\APACjournalVolNumPages{The Journal of physiology}{160}{1}{106}.
\PrintBackRefs{\CurrentBib}

\bibitem [\protect \citeauthoryear {%
Hubel%
\ \BBA {} Wiesel%
}{%
Hubel%
\ \BBA {} Wiesel%
}{%
{\protect \APACyear {1968}}%
}]{%
hubel1968receptive}
\APACinsertmetastar {%
hubel1968receptive}%
\begin{APACrefauthors}%
Hubel, D\BPBI H.%
\BCBT {}\ \BBA {} Wiesel, T\BPBI N.%
\end{APACrefauthors}%
\unskip\
\newblock
\APACrefYearMonthDay{1968}{}{}.
\newblock
{\BBOQ}\APACrefatitle {Receptive fields and functional architecture of monkey striate cortex} {Receptive fields and functional architecture of monkey striate cortex}.{\BBCQ}
\newblock
\APACjournalVolNumPages{The Journal of physiology}{195}{1}{215--243}.
\PrintBackRefs{\CurrentBib}

\bibitem [\protect \citeauthoryear {%
Huh%
, Cheung%
, Wang%
\BCBL {}\ \BBA {} Isola%
}{%
Huh%
\ \protect \BOthers {.}}{%
{\protect \APACyear {2024}}%
}]{%
huh2024platonic}
\APACinsertmetastar {%
huh2024platonic}%
\begin{APACrefauthors}%
Huh, M.%
, Cheung, B.%
, Wang, T.%
\BCBL {}\ \BBA {} Isola, P.%
\end{APACrefauthors}%
\unskip\
\newblock
\APACrefYearMonthDay{2024}{}{}.
\newblock
{\BBOQ}\APACrefatitle {The platonic representation hypothesis} {The platonic representation hypothesis}.{\BBCQ}
\newblock
\APACjournalVolNumPages{arXiv preprint arXiv:2405.07987}{}{}{}.
\PrintBackRefs{\CurrentBib}

\bibitem [\protect \citeauthoryear {%
Huth%
, Nishimoto%
, Vu%
\BCBL {}\ \BBA {} Gallant%
}{%
Huth%
\ \protect \BOthers {.}}{%
{\protect \APACyear {2012}}%
}]{%
huth2012continuous}
\APACinsertmetastar {%
huth2012continuous}%
\begin{APACrefauthors}%
Huth, A\BPBI G.%
, Nishimoto, S.%
, Vu, A\BPBI T.%
\BCBL {}\ \BBA {} Gallant, J\BPBI L.%
\end{APACrefauthors}%
\unskip\
\newblock
\APACrefYearMonthDay{2012}{}{}.
\newblock
{\BBOQ}\APACrefatitle {A continuous semantic space describes the representation of thousands of object and action categories across the human brain} {A continuous semantic space describes the representation of thousands of object and action categories across the human brain}.{\BBCQ}
\newblock
\APACjournalVolNumPages{Neuron}{76}{6}{1210--1224}.
\PrintBackRefs{\CurrentBib}

\bibitem [\protect \citeauthoryear {%
Ivanova%
\ \protect \BOthers {.}}{%
Ivanova%
\ \protect \BOthers {.}}{%
{\protect \APACyear {2022}}%
}]{%
ivanova2022beyond}
\APACinsertmetastar {%
ivanova2022beyond}%
\begin{APACrefauthors}%
Ivanova, A\BPBI A.%
, Schrimpf, M.%
, Anzellotti, S.%
, Zaslavsky, N.%
, Fedorenko, E.%
\BCBL {}\ \BBA {} Isik, L.%
\end{APACrefauthors}%
\unskip\
\newblock
\APACrefYearMonthDay{2022}{}{}.
\newblock
{\BBOQ}\APACrefatitle {Beyond linear regression: mapping models in cognitive neuroscience should align with research goals} {Beyond linear regression: mapping models in cognitive neuroscience should align with research goals}.{\BBCQ}
\newblock
\APACjournalVolNumPages{arXiv preprint arXiv:2208.10668}{}{}{}.
\PrintBackRefs{\CurrentBib}

\bibitem [\protect \citeauthoryear {%
Jaderberg%
, Simonyan%
, Zisserman%
\BCBL {}\ \protect \BOthers {.}}{%
Jaderberg%
\ \protect \BOthers {.}}{%
{\protect \APACyear {2015}}%
}]{%
jaderberg2015spatial}
\APACinsertmetastar {%
jaderberg2015spatial}%
\begin{APACrefauthors}%
Jaderberg, M.%
, Simonyan, K.%
, Zisserman, A.%
\BCBL {}\ \BOthersPeriod {.}\end{APACrefauthors}%
\unskip\
\newblock
\APACrefYearMonthDay{2015}{}{}.
\newblock
{\BBOQ}\APACrefatitle {Spatial transformer networks} {Spatial transformer networks}.{\BBCQ}
\newblock
\APACjournalVolNumPages{Advances in neural information processing systems}{28}{}{}.
\PrintBackRefs{\CurrentBib}

\bibitem [\protect \citeauthoryear {%
Khaligh-Razavi%
\ \BBA {} Kriegeskorte%
}{%
Khaligh-Razavi%
\ \BBA {} Kriegeskorte%
}{%
{\protect \APACyear {2014}}%
}]{%
khaligh2014deep}
\APACinsertmetastar {%
khaligh2014deep}%
\begin{APACrefauthors}%
Khaligh-Razavi, S\BHBI M.%
\BCBT {}\ \BBA {} Kriegeskorte, N.%
\end{APACrefauthors}%
\unskip\
\newblock
\APACrefYearMonthDay{2014}{}{}.
\newblock
{\BBOQ}\APACrefatitle {Deep supervised, but not unsupervised, models may explain IT cortical representation} {Deep supervised, but not unsupervised, models may explain it cortical representation}.{\BBCQ}
\newblock
\APACjournalVolNumPages{PLoS computational biology}{10}{11}{e1003915}.
\PrintBackRefs{\CurrentBib}

\bibitem [\protect \citeauthoryear {%
Khosla%
, Jamison%
, Kuceyeski%
\BCBL {}\ \BBA {} Sabuncu%
}{%
Khosla%
\ \protect \BOthers {.}}{%
{\protect \APACyear {2022}}%
}]{%
khosla2022characterizing}
\APACinsertmetastar {%
khosla2022characterizing}%
\begin{APACrefauthors}%
Khosla, M.%
, Jamison, K.%
, Kuceyeski, A.%
\BCBL {}\ \BBA {} Sabuncu, M.%
\end{APACrefauthors}%
\unskip\
\newblock
\APACrefYearMonthDay{2022}{}{}.
\newblock
{\BBOQ}\APACrefatitle {Characterizing the ventral visual stream with response-optimized neural encoding models} {Characterizing the ventral visual stream with response-optimized neural encoding models}.{\BBCQ}
\newblock
\APACjournalVolNumPages{Advances in Neural Information Processing Systems}{35}{}{9389--9402}.
\PrintBackRefs{\CurrentBib}

\bibitem [\protect \citeauthoryear {%
Khosla%
\ \BBA {} Wehbe%
}{%
Khosla%
\ \BBA {} Wehbe%
}{%
{\protect \APACyear {2022}}%
}]{%
khosla2022high}
\APACinsertmetastar {%
khosla2022high}%
\begin{APACrefauthors}%
Khosla, M.%
\BCBT {}\ \BBA {} Wehbe, L.%
\end{APACrefauthors}%
\unskip\
\newblock
\APACrefYearMonthDay{2022}{}{}.
\newblock
{\BBOQ}\APACrefatitle {High-level visual areas act like domain-general filters with strong selectivity and functional specialization} {High-level visual areas act like domain-general filters with strong selectivity and functional specialization}.{\BBCQ}
\newblock
\APACjournalVolNumPages{bioRxiv}{}{}{2022--03}.
\PrintBackRefs{\CurrentBib}

\bibitem [\protect \citeauthoryear {%
Klindt%
, Ecker%
, Euler%
\BCBL {}\ \BBA {} Bethge%
}{%
Klindt%
\ \protect \BOthers {.}}{%
{\protect \APACyear {2017}}%
}]{%
klindt2017neural}
\APACinsertmetastar {%
klindt2017neural}%
\begin{APACrefauthors}%
Klindt, D.%
, Ecker, A.%
, Euler, T.%
\BCBL {}\ \BBA {} Bethge, M.%
\end{APACrefauthors}%
\unskip\
\newblock
\APACrefYearMonthDay{2017}{}{}.
\newblock
{\BBOQ}\APACrefatitle {Neural system identification for large 579 populations separating “what” and “where.”} {Neural system identification for large 579 populations separating “what” and “where.”}.{\BBCQ}
\newblock
\APACjournalVolNumPages{Advances in Neural Information Processing 580 Systems}{}{}{}.
\PrintBackRefs{\CurrentBib}

\bibitem [\protect \citeauthoryear {%
Kobatake%
\ \BBA {} Tanaka%
}{%
Kobatake%
\ \BBA {} Tanaka%
}{%
{\protect \APACyear {1994}}%
}]{%
kobatake1994neuronal}
\APACinsertmetastar {%
kobatake1994neuronal}%
\begin{APACrefauthors}%
Kobatake, E.%
\BCBT {}\ \BBA {} Tanaka, K.%
\end{APACrefauthors}%
\unskip\
\newblock
\APACrefYearMonthDay{1994}{}{}.
\newblock
{\BBOQ}\APACrefatitle {Neuronal selectivities to complex object features in the ventral visual pathway of the macaque cerebral cortex} {Neuronal selectivities to complex object features in the ventral visual pathway of the macaque cerebral cortex}.{\BBCQ}
\newblock
\APACjournalVolNumPages{Journal of neurophysiology}{71}{3}{856--867}.
\PrintBackRefs{\CurrentBib}

\bibitem [\protect \citeauthoryear {%
Kobatake%
, Wang%
\BCBL {}\ \BBA {} Tanaka%
}{%
Kobatake%
\ \protect \BOthers {.}}{%
{\protect \APACyear {1998}}%
}]{%
kobatake1998effects}
\APACinsertmetastar {%
kobatake1998effects}%
\begin{APACrefauthors}%
Kobatake, E.%
, Wang, G.%
\BCBL {}\ \BBA {} Tanaka, K.%
\end{APACrefauthors}%
\unskip\
\newblock
\APACrefYearMonthDay{1998}{}{}.
\newblock
{\BBOQ}\APACrefatitle {Effects of shape-discrimination training on the selectivity of inferotemporal cells in adult monkeys} {Effects of shape-discrimination training on the selectivity of inferotemporal cells in adult monkeys}.{\BBCQ}
\newblock
\APACjournalVolNumPages{Journal of neurophysiology}{80}{1}{324--330}.
\PrintBackRefs{\CurrentBib}

\bibitem [\protect \citeauthoryear {%
Kriegeskorte%
\ \protect \BOthers {.}}{%
Kriegeskorte%
\ \protect \BOthers {.}}{%
{\protect \APACyear {2008}}%
}]{%
kriegeskorte2008matching}
\APACinsertmetastar {%
kriegeskorte2008matching}%
\begin{APACrefauthors}%
Kriegeskorte, N.%
, Mur, M.%
, Ruff, D\BPBI A.%
, Kiani, R.%
, Bodurka, J.%
, Esteky, H.%
\BDBL {}Bandettini, P\BPBI A.%
\end{APACrefauthors}%
\unskip\
\newblock
\APACrefYearMonthDay{2008}{}{}.
\newblock
{\BBOQ}\APACrefatitle {Matching categorical object representations in inferior temporal cortex of man and monkey} {Matching categorical object representations in inferior temporal cortex of man and monkey}.{\BBCQ}
\newblock
\APACjournalVolNumPages{Neuron}{60}{6}{1126--1141}.
\PrintBackRefs{\CurrentBib}

\bibitem [\protect \citeauthoryear {%
Krizhevsky%
, Sutskever%
\BCBL {}\ \BBA {} Hinton%
}{%
Krizhevsky%
\ \protect \BOthers {.}}{%
{\protect \APACyear {2012}}%
}]{%
krizhevsky2012imagenet}
\APACinsertmetastar {%
krizhevsky2012imagenet}%
\begin{APACrefauthors}%
Krizhevsky, A.%
, Sutskever, I.%
\BCBL {}\ \BBA {} Hinton, G\BPBI E.%
\end{APACrefauthors}%
\unskip\
\newblock
\APACrefYearMonthDay{2012}{}{}.
\newblock
{\BBOQ}\APACrefatitle {Imagenet classification with deep convolutional neural networks} {Imagenet classification with deep convolutional neural networks}.{\BBCQ}
\newblock
\APACjournalVolNumPages{Advances in neural information processing systems}{25}{}{}.
\PrintBackRefs{\CurrentBib}

\bibitem [\protect \citeauthoryear {%
Lin%
\ \protect \BOthers {.}}{%
Lin%
\ \protect \BOthers {.}}{%
{\protect \APACyear {2014}}%
}]{%
lin2014microsoft}
\APACinsertmetastar {%
lin2014microsoft}%
\begin{APACrefauthors}%
Lin, T\BHBI Y.%
, Maire, M.%
, Belongie, S.%
, Hays, J.%
, Perona, P.%
, Ramanan, D.%
\BDBL {}Zitnick, C\BPBI L.%
\end{APACrefauthors}%
\unskip\
\newblock
\APACrefYearMonthDay{2014}{}{}.
\newblock
{\BBOQ}\APACrefatitle {Microsoft coco: Common objects in context} {Microsoft coco: Common objects in context}.{\BBCQ}
\newblock
\BIn{} \APACrefbtitle {Computer Vision--ECCV 2014: 13th European Conference, Zurich, Switzerland, September 6-12, 2014, Proceedings, Part V 13} {Computer vision--eccv 2014: 13th european conference, zurich, switzerland, september 6-12, 2014, proceedings, part v 13}\ (\BPGS\ 740--755).
\PrintBackRefs{\CurrentBib}

\bibitem [\protect \citeauthoryear {%
Liu%
\ \protect \BOthers {.}}{%
Liu%
\ \protect \BOthers {.}}{%
{\protect \APACyear {2022}}%
}]{%
liu2022convnet}
\APACinsertmetastar {%
liu2022convnet}%
\begin{APACrefauthors}%
Liu, Z.%
, Mao, H.%
, Wu, C\BHBI Y.%
, Feichtenhofer, C.%
, Darrell, T.%
\BCBL {}\ \BBA {} Xie, S.%
\end{APACrefauthors}%
\unskip\
\newblock
\APACrefYearMonthDay{2022}{}{}.
\newblock
{\BBOQ}\APACrefatitle {A convnet for the 2020s} {A convnet for the 2020s}.{\BBCQ}
\newblock
\BIn{} \APACrefbtitle {Proceedings of the IEEE/CVF conference on computer vision and pattern recognition} {Proceedings of the ieee/cvf conference on computer vision and pattern recognition}\ (\BPGS\ 11976--11986).
\PrintBackRefs{\CurrentBib}

\bibitem [\protect \citeauthoryear {%
Livingstone%
\ \BBA {} Hubel%
}{%
Livingstone%
\ \BBA {} Hubel%
}{%
{\protect \APACyear {1984}}%
}]{%
livingstone1984anatomy}
\APACinsertmetastar {%
livingstone1984anatomy}%
\begin{APACrefauthors}%
Livingstone, M\BPBI S.%
\BCBT {}\ \BBA {} Hubel, D\BPBI H.%
\end{APACrefauthors}%
\unskip\
\newblock
\APACrefYearMonthDay{1984}{}{}.
\newblock
{\BBOQ}\APACrefatitle {Anatomy and physiology of a color system in the primate visual cortex} {Anatomy and physiology of a color system in the primate visual cortex}.{\BBCQ}
\newblock
\APACjournalVolNumPages{Journal of Neuroscience}{4}{1}{309--356}.
\PrintBackRefs{\CurrentBib}

\bibitem [\protect \citeauthoryear {%
Lurz%
\ \protect \BOthers {.}}{%
Lurz%
\ \protect \BOthers {.}}{%
{\protect \APACyear {2020}}%
}]{%
lurz2020generalization}
\APACinsertmetastar {%
lurz2020generalization}%
\begin{APACrefauthors}%
Lurz, K\BHBI K.%
, Bashiri, M.%
, Willeke, K.%
, Jagadish, A\BPBI K.%
, Wang, E.%
, Walker, E\BPBI Y.%
\BDBL {}others%
\end{APACrefauthors}%
\unskip\
\newblock
\APACrefYearMonthDay{2020}{}{}.
\newblock
{\BBOQ}\APACrefatitle {Generalization in data-driven models of primary visual cortex} {Generalization in data-driven models of primary visual cortex}.{\BBCQ}
\newblock
\APACjournalVolNumPages{BioRxiv}{}{}{2020--10}.
\PrintBackRefs{\CurrentBib}

\bibitem [\protect \citeauthoryear {%
Miyashita%
}{%
Miyashita%
}{%
{\protect \APACyear {1988}}%
}]{%
miyashita1988neuronal}
\APACinsertmetastar {%
miyashita1988neuronal}%
\begin{APACrefauthors}%
Miyashita, Y.%
\end{APACrefauthors}%
\unskip\
\newblock
\APACrefYearMonthDay{1988}{}{}.
\newblock
{\BBOQ}\APACrefatitle {Neuronal correlate of visual associative long-term memory in the primate temporal cortex} {Neuronal correlate of visual associative long-term memory in the primate temporal cortex}.{\BBCQ}
\newblock
\APACjournalVolNumPages{Nature}{335}{6193}{817--820}.
\PrintBackRefs{\CurrentBib}

\bibitem [\protect \citeauthoryear {%
Moran%
\ \BBA {} Desimone%
}{%
Moran%
\ \BBA {} Desimone%
}{%
{\protect \APACyear {1985}}%
}]{%
moran1985selective}
\APACinsertmetastar {%
moran1985selective}%
\begin{APACrefauthors}%
Moran, J.%
\BCBT {}\ \BBA {} Desimone, R.%
\end{APACrefauthors}%
\unskip\
\newblock
\APACrefYearMonthDay{1985}{}{}.
\newblock
{\BBOQ}\APACrefatitle {Selective attention gates visual processing in the extrastriate cortex} {Selective attention gates visual processing in the extrastriate cortex}.{\BBCQ}
\newblock
\APACjournalVolNumPages{Science}{229}{4715}{782--784}.
\PrintBackRefs{\CurrentBib}

\bibitem [\protect \citeauthoryear {%
Pasupathy%
\ \BBA {} Connor%
}{%
Pasupathy%
\ \BBA {} Connor%
}{%
{\protect \APACyear {1999}}%
}]{%
pasupathy1999responses}
\APACinsertmetastar {%
pasupathy1999responses}%
\begin{APACrefauthors}%
Pasupathy, A.%
\BCBT {}\ \BBA {} Connor, C\BPBI E.%
\end{APACrefauthors}%
\unskip\
\newblock
\APACrefYearMonthDay{1999}{}{}.
\newblock
{\BBOQ}\APACrefatitle {Responses to contour features in macaque area V4} {Responses to contour features in macaque area v4}.{\BBCQ}
\newblock
\APACjournalVolNumPages{Journal of neurophysiology}{82}{5}{2490--2502}.
\PrintBackRefs{\CurrentBib}

\bibitem [\protect \citeauthoryear {%
Pasupathy%
\ \BBA {} Connor%
}{%
Pasupathy%
\ \BBA {} Connor%
}{%
{\protect \APACyear {2001}}%
}]{%
pasupathy2001shape}
\APACinsertmetastar {%
pasupathy2001shape}%
\begin{APACrefauthors}%
Pasupathy, A.%
\BCBT {}\ \BBA {} Connor, C\BPBI E.%
\end{APACrefauthors}%
\unskip\
\newblock
\APACrefYearMonthDay{2001}{}{}.
\newblock
{\BBOQ}\APACrefatitle {Shape representation in area V4: position-specific tuning for boundary conformation} {Shape representation in area v4: position-specific tuning for boundary conformation}.{\BBCQ}
\newblock
\APACjournalVolNumPages{Journal of neurophysiology}{}{}{}.
\PrintBackRefs{\CurrentBib}

\bibitem [\protect \citeauthoryear {%
Pasupathy%
\ \BBA {} Connor%
}{%
Pasupathy%
\ \BBA {} Connor%
}{%
{\protect \APACyear {2002}}%
}]{%
pasupathy2002population}
\APACinsertmetastar {%
pasupathy2002population}%
\begin{APACrefauthors}%
Pasupathy, A.%
\BCBT {}\ \BBA {} Connor, C\BPBI E.%
\end{APACrefauthors}%
\unskip\
\newblock
\APACrefYearMonthDay{2002}{}{}.
\newblock
{\BBOQ}\APACrefatitle {Population coding of shape in area V4} {Population coding of shape in area v4}.{\BBCQ}
\newblock
\APACjournalVolNumPages{Nature neuroscience}{5}{12}{1332--1338}.
\PrintBackRefs{\CurrentBib}

\bibitem [\protect \citeauthoryear {%
Radford%
\ \protect \BOthers {.}}{%
Radford%
\ \protect \BOthers {.}}{%
{\protect \APACyear {2021}}%
}]{%
radford2021learning}
\APACinsertmetastar {%
radford2021learning}%
\begin{APACrefauthors}%
Radford, A.%
, Kim, J\BPBI W.%
, Hallacy, C.%
, Ramesh, A.%
, Goh, G.%
, Agarwal, S.%
\BDBL {}others%
\end{APACrefauthors}%
\unskip\
\newblock
\APACrefYearMonthDay{2021}{}{}.
\newblock
{\BBOQ}\APACrefatitle {Learning transferable visual models from natural language supervision} {Learning transferable visual models from natural language supervision}.{\BBCQ}
\newblock
\BIn{} \APACrefbtitle {International conference on machine learning} {International conference on machine learning}\ (\BPGS\ 8748--8763).
\PrintBackRefs{\CurrentBib}

\bibitem [\protect \citeauthoryear {%
Rust%
\ \BBA {} DiCarlo%
}{%
Rust%
\ \BBA {} DiCarlo%
}{%
{\protect \APACyear {2010}}%
}]{%
rust2010selectivity}
\APACinsertmetastar {%
rust2010selectivity}%
\begin{APACrefauthors}%
Rust, N\BPBI C.%
\BCBT {}\ \BBA {} DiCarlo, J\BPBI J.%
\end{APACrefauthors}%
\unskip\
\newblock
\APACrefYearMonthDay{2010}{}{}.
\newblock
{\BBOQ}\APACrefatitle {Selectivity and tolerance (“invariance”) both increase as visual information propagates from cortical area V4 to IT} {Selectivity and tolerance (“invariance”) both increase as visual information propagates from cortical area v4 to it}.{\BBCQ}
\newblock
\APACjournalVolNumPages{Journal of Neuroscience}{30}{39}{12978--12995}.
\PrintBackRefs{\CurrentBib}

\bibitem [\protect \citeauthoryear {%
Safarani%
\ \protect \BOthers {.}}{%
Safarani%
\ \protect \BOthers {.}}{%
{\protect \APACyear {2021}}%
}]{%
safarani2021towards}
\APACinsertmetastar {%
safarani2021towards}%
\begin{APACrefauthors}%
Safarani, S.%
, Nix, A.%
, Willeke, K.%
, Cadena, S.%
, Restivo, K.%
, Denfield, G.%
\BDBL {}Sinz, F.%
\end{APACrefauthors}%
\unskip\
\newblock
\APACrefYearMonthDay{2021}{}{}.
\newblock
{\BBOQ}\APACrefatitle {Towards robust vision by multi-task learning on monkey visual cortex} {Towards robust vision by multi-task learning on monkey visual cortex}.{\BBCQ}
\newblock
\APACjournalVolNumPages{Advances in Neural Information Processing Systems}{34}{}{739--751}.
\PrintBackRefs{\CurrentBib}

\bibitem [\protect \citeauthoryear {%
Sceniak%
, Ringach%
, Hawken%
\BCBL {}\ \BBA {} Shapley%
}{%
Sceniak%
\ \protect \BOthers {.}}{%
{\protect \APACyear {1999}}%
}]{%
sceniak1999contrast}
\APACinsertmetastar {%
sceniak1999contrast}%
\begin{APACrefauthors}%
Sceniak, M\BPBI P.%
, Ringach, D\BPBI L.%
, Hawken, M\BPBI J.%
\BCBL {}\ \BBA {} Shapley, R.%
\end{APACrefauthors}%
\unskip\
\newblock
\APACrefYearMonthDay{1999}{}{}.
\newblock
{\BBOQ}\APACrefatitle {Contrast's effect on spatial summation by macaque V1 neurons} {Contrast's effect on spatial summation by macaque v1 neurons}.{\BBCQ}
\newblock
\APACjournalVolNumPages{Nature neuroscience}{2}{8}{733--739}.
\PrintBackRefs{\CurrentBib}

\bibitem [\protect \citeauthoryear {%
Schrimpf%
\ \protect \BOthers {.}}{%
Schrimpf%
\ \protect \BOthers {.}}{%
{\protect \APACyear {2020}}%
}]{%
schrimpf2020integrative}
\APACinsertmetastar {%
schrimpf2020integrative}%
\begin{APACrefauthors}%
Schrimpf, M.%
, Kubilius, J.%
, Lee, M\BPBI J.%
, Murty, N\BPBI A\BPBI R.%
, Ajemian, R.%
\BCBL {}\ \BBA {} DiCarlo, J\BPBI J.%
\end{APACrefauthors}%
\unskip\
\newblock
\APACrefYearMonthDay{2020}{}{}.
\newblock
{\BBOQ}\APACrefatitle {Integrative benchmarking to advance neurally mechanistic models of human intelligence} {Integrative benchmarking to advance neurally mechanistic models of human intelligence}.{\BBCQ}
\newblock
\APACjournalVolNumPages{Neuron}{108}{3}{413--423}.
\PrintBackRefs{\CurrentBib}

\bibitem [\protect \citeauthoryear {%
Schwartz%
, Toneva%
\BCBL {}\ \BBA {} Wehbe%
}{%
Schwartz%
\ \protect \BOthers {.}}{%
{\protect \APACyear {2019}}%
}]{%
schwartz2019inducing}
\APACinsertmetastar {%
schwartz2019inducing}%
\begin{APACrefauthors}%
Schwartz, D.%
, Toneva, M.%
\BCBL {}\ \BBA {} Wehbe, L.%
\end{APACrefauthors}%
\unskip\
\newblock
\APACrefYearMonthDay{2019}{}{}.
\newblock
{\BBOQ}\APACrefatitle {Inducing brain-relevant bias in natural language processing models} {Inducing brain-relevant bias in natural language processing models}.{\BBCQ}
\newblock
\APACjournalVolNumPages{Advances in neural information processing systems}{32}{}{}.
\PrintBackRefs{\CurrentBib}

\bibitem [\protect \citeauthoryear {%
Seeliger%
\ \protect \BOthers {.}}{%
Seeliger%
\ \protect \BOthers {.}}{%
{\protect \APACyear {2021}}%
}]{%
seeliger2021end}
\APACinsertmetastar {%
seeliger2021end}%
\begin{APACrefauthors}%
Seeliger, K.%
, Ambrogioni, L.%
, G{\"u}{\c{c}}l{\"u}t{\"u}rk, Y.%
, van~den Bulk, L\BPBI M.%
, G{\"u}{\c{c}}l{\"u}, U.%
\BCBL {}\ \BBA {} van Gerven, M\BPBI A.%
\end{APACrefauthors}%
\unskip\
\newblock
\APACrefYearMonthDay{2021}{}{}.
\newblock
{\BBOQ}\APACrefatitle {End-to-end neural system identification with neural information flow} {End-to-end neural system identification with neural information flow}.{\BBCQ}
\newblock
\APACjournalVolNumPages{PLOS Computational Biology}{17}{2}{e1008558}.
\PrintBackRefs{\CurrentBib}

\bibitem [\protect \citeauthoryear {%
Shen%
, Conwell%
\BCBL {}\ \BBA {} Bonner%
}{%
Shen%
\ \protect \BOthers {.}}{%
{\protect \APACyear {{\protect \bibnodate {}}}}%
}]{%
shenhigh}
\APACinsertmetastar {%
shenhigh}%
\begin{APACrefauthors}%
Shen, T.%
, Conwell, C.%
\BCBL {}\ \BBA {} Bonner, M\BPBI F.%
\end{APACrefauthors}%
\unskip\
\newblock
\APACrefYearMonthDay{{\protect \bibnodate {}}}{}{}.
\newblock
{\BBOQ}\APACrefatitle {High-dimensional alignment of neural networks and visual cortex} {High-dimensional alignment of neural networks and visual cortex}.{\BBCQ}
\newblock

\PrintBackRefs{\CurrentBib}

\bibitem [\protect \citeauthoryear {%
Shoham%
, Broday-Dvir%
, Malach%
\BCBL {}\ \BBA {} Yovel%
}{%
Shoham%
\ \protect \BOthers {.}}{%
{\protect \APACyear {2024}}%
}]{%
shoham2024organization}
\APACinsertmetastar {%
shoham2024organization}%
\begin{APACrefauthors}%
Shoham, A.%
, Broday-Dvir, R.%
, Malach, R.%
\BCBL {}\ \BBA {} Yovel, G.%
\end{APACrefauthors}%
\unskip\
\newblock
\APACrefYearMonthDay{2024}{}{}.
\newblock
{\BBOQ}\APACrefatitle {The organization of high-level visual cortex is aligned with visual rather than abstract linguistic information} {The organization of high-level visual cortex is aligned with visual rather than abstract linguistic information}.{\BBCQ}
\newblock
\APACjournalVolNumPages{bioRxiv}{}{}{2024--11}.
\PrintBackRefs{\CurrentBib}

\bibitem [\protect \citeauthoryear {%
Song%
, Tan%
, Qin%
, Lu%
\BCBL {}\ \BBA {} Liu%
}{%
Song%
\ \protect \BOthers {.}}{%
{\protect \APACyear {2020}}%
}]{%
song2020mpnet}
\APACinsertmetastar {%
song2020mpnet}%
\begin{APACrefauthors}%
Song, K.%
, Tan, X.%
, Qin, T.%
, Lu, J.%
\BCBL {}\ \BBA {} Liu, T\BHBI Y.%
\end{APACrefauthors}%
\unskip\
\newblock
\APACrefYearMonthDay{2020}{}{}.
\newblock
{\BBOQ}\APACrefatitle {Mpnet: Masked and permuted pre-training for language understanding} {Mpnet: Masked and permuted pre-training for language understanding}.{\BBCQ}
\newblock
\APACjournalVolNumPages{Advances in neural information processing systems}{33}{}{16857--16867}.
\PrintBackRefs{\CurrentBib}

\bibitem [\protect \citeauthoryear {%
Storrs%
, Kietzmann%
, Walther%
, Mehrer%
\BCBL {}\ \BBA {} Kriegeskorte%
}{%
Storrs%
\ \protect \BOthers {.}}{%
{\protect \APACyear {2021}}%
}]{%
storrs2021diverse}
\APACinsertmetastar {%
storrs2021diverse}%
\begin{APACrefauthors}%
Storrs, K\BPBI R.%
, Kietzmann, T\BPBI C.%
, Walther, A.%
, Mehrer, J.%
\BCBL {}\ \BBA {} Kriegeskorte, N.%
\end{APACrefauthors}%
\unskip\
\newblock
\APACrefYearMonthDay{2021}{}{}.
\newblock
{\BBOQ}\APACrefatitle {Diverse deep neural networks all predict human inferior temporal cortex well, after training and fitting} {Diverse deep neural networks all predict human inferior temporal cortex well, after training and fitting}.{\BBCQ}
\newblock
\APACjournalVolNumPages{Journal of cognitive neuroscience}{33}{10}{2044--2064}.
\PrintBackRefs{\CurrentBib}

\bibitem [\protect \citeauthoryear {%
St-Yves%
, Allen%
, Wu%
, Kay%
\BCBL {}\ \BBA {} Naselaris%
}{%
St-Yves%
\ \protect \BOthers {.}}{%
{\protect \APACyear {2023}}%
}]{%
st2023brain}
\APACinsertmetastar {%
st2023brain}%
\begin{APACrefauthors}%
St-Yves, G.%
, Allen, E\BPBI J.%
, Wu, Y.%
, Kay, K.%
\BCBL {}\ \BBA {} Naselaris, T.%
\end{APACrefauthors}%
\unskip\
\newblock
\APACrefYearMonthDay{2023}{}{}.
\newblock
{\BBOQ}\APACrefatitle {Brain-optimized deep neural network models of human visual areas learn non-hierarchical representations} {Brain-optimized deep neural network models of human visual areas learn non-hierarchical representations}.{\BBCQ}
\newblock
\APACjournalVolNumPages{Nature communications}{14}{1}{3329}.
\PrintBackRefs{\CurrentBib}

\bibitem [\protect \citeauthoryear {%
Tanaka%
, Saito%
, Fukada%
\BCBL {}\ \BBA {} Moriya%
}{%
Tanaka%
\ \protect \BOthers {.}}{%
{\protect \APACyear {1991}}%
}]{%
tanaka1991coding}
\APACinsertmetastar {%
tanaka1991coding}%
\begin{APACrefauthors}%
Tanaka, K.%
, Saito, H\BHBI a.%
, Fukada, Y.%
\BCBL {}\ \BBA {} Moriya, M.%
\end{APACrefauthors}%
\unskip\
\newblock
\APACrefYearMonthDay{1991}{}{}.
\newblock
{\BBOQ}\APACrefatitle {Coding visual images of objects in the inferotemporal cortex of the macaque monkey} {Coding visual images of objects in the inferotemporal cortex of the macaque monkey}.{\BBCQ}
\newblock
\APACjournalVolNumPages{Journal of neurophysiology}{66}{1}{170--189}.
\PrintBackRefs{\CurrentBib}

\bibitem [\protect \citeauthoryear {%
Tang%
, Du%
, Vo%
, Lal%
\BCBL {}\ \BBA {} Huth%
}{%
Tang%
\ \protect \BOthers {.}}{%
{\protect \APACyear {2024}}%
}]{%
tang2024brain}
\APACinsertmetastar {%
tang2024brain}%
\begin{APACrefauthors}%
Tang, J.%
, Du, M.%
, Vo, V.%
, Lal, V.%
\BCBL {}\ \BBA {} Huth, A.%
\end{APACrefauthors}%
\unskip\
\newblock
\APACrefYearMonthDay{2024}{}{}.
\newblock
{\BBOQ}\APACrefatitle {Brain encoding models based on multimodal transformers can transfer across language and vision} {Brain encoding models based on multimodal transformers can transfer across language and vision}.{\BBCQ}
\newblock
\APACjournalVolNumPages{Advances in Neural Information Processing Systems}{36}{}{}.
\PrintBackRefs{\CurrentBib}

\bibitem [\protect \citeauthoryear {%
Tsunoda%
, Yamane%
, Nishizaki%
\BCBL {}\ \BBA {} Tanifuji%
}{%
Tsunoda%
\ \protect \BOthers {.}}{%
{\protect \APACyear {2001}}%
}]{%
tsunoda2001complex}
\APACinsertmetastar {%
tsunoda2001complex}%
\begin{APACrefauthors}%
Tsunoda, K.%
, Yamane, Y.%
, Nishizaki, M.%
\BCBL {}\ \BBA {} Tanifuji, M.%
\end{APACrefauthors}%
\unskip\
\newblock
\APACrefYearMonthDay{2001}{}{}.
\newblock
{\BBOQ}\APACrefatitle {Complex objects are represented in macaque inferotemporal cortex by the combination of feature columns} {Complex objects are represented in macaque inferotemporal cortex by the combination of feature columns}.{\BBCQ}
\newblock
\APACjournalVolNumPages{Nature neuroscience}{4}{8}{832--838}.
\PrintBackRefs{\CurrentBib}

\bibitem [\protect \citeauthoryear {%
Walker%
\ \protect \BOthers {.}}{%
Walker%
\ \protect \BOthers {.}}{%
{\protect \APACyear {2019}}%
}]{%
walker2019inception}
\APACinsertmetastar {%
walker2019inception}%
\begin{APACrefauthors}%
Walker, E\BPBI Y.%
, Sinz, F\BPBI H.%
, Cobos, E.%
, Muhammad, T.%
, Froudarakis, E.%
, Fahey, P\BPBI G.%
\BDBL {}Tolias, A\BPBI S.%
\end{APACrefauthors}%
\unskip\
\newblock
\APACrefYearMonthDay{2019}{}{}.
\newblock
{\BBOQ}\APACrefatitle {Inception loops discover what excites neurons most using deep predictive models} {Inception loops discover what excites neurons most using deep predictive models}.{\BBCQ}
\newblock
\APACjournalVolNumPages{Nature neuroscience}{22}{12}{2060--2065}.
\PrintBackRefs{\CurrentBib}

\bibitem [\protect \citeauthoryear {%
Wang%
, Kay%
, Naselaris%
, Tarr%
\BCBL {}\ \BBA {} Wehbe%
}{%
Wang%
\ \protect \BOthers {.}}{%
{\protect \APACyear {2022}}%
}]{%
wang2022incorporating}
\APACinsertmetastar {%
wang2022incorporating}%
\begin{APACrefauthors}%
Wang, A\BPBI Y.%
, Kay, K.%
, Naselaris, T.%
, Tarr, M\BPBI J.%
\BCBL {}\ \BBA {} Wehbe, L.%
\end{APACrefauthors}%
\unskip\
\newblock
\APACrefYearMonthDay{2022}{}{}.
\newblock
{\BBOQ}\APACrefatitle {Incorporating natural language into vision models improves prediction and understanding of higher visual cortex} {Incorporating natural language into vision models improves prediction and understanding of higher visual cortex}.{\BBCQ}
\newblock
\APACjournalVolNumPages{BioRxiv}{}{}{2022--09}.
\PrintBackRefs{\CurrentBib}

\bibitem [\protect \citeauthoryear {%
Weiler%
\ \BBA {} Cesa%
}{%
Weiler%
\ \BBA {} Cesa%
}{%
{\protect \APACyear {2019}}%
}]{%
weiler2019general}
\APACinsertmetastar {%
weiler2019general}%
\begin{APACrefauthors}%
Weiler, M.%
\BCBT {}\ \BBA {} Cesa, G.%
\end{APACrefauthors}%
\unskip\
\newblock
\APACrefYearMonthDay{2019}{}{}.
\newblock
{\BBOQ}\APACrefatitle {General e (2)-equivariant steerable cnns} {General e (2)-equivariant steerable cnns}.{\BBCQ}
\newblock
\APACjournalVolNumPages{Advances in neural information processing systems}{32}{}{}.
\PrintBackRefs{\CurrentBib}

\bibitem [\protect \citeauthoryear {%
Wen%
\ \protect \BOthers {.}}{%
Wen%
\ \protect \BOthers {.}}{%
{\protect \APACyear {2018}}%
}]{%
wen2018neural}
\APACinsertmetastar {%
wen2018neural}%
\begin{APACrefauthors}%
Wen, H.%
, Shi, J.%
, Zhang, Y.%
, Lu, K\BHBI H.%
, Cao, J.%
\BCBL {}\ \BBA {} Liu, Z.%
\end{APACrefauthors}%
\unskip\
\newblock
\APACrefYearMonthDay{2018}{}{}.
\newblock
{\BBOQ}\APACrefatitle {Neural encoding and decoding with deep learning for dynamic natural vision} {Neural encoding and decoding with deep learning for dynamic natural vision}.{\BBCQ}
\newblock
\APACjournalVolNumPages{Cerebral cortex}{28}{12}{4136--4160}.
\PrintBackRefs{\CurrentBib}

\bibitem [\protect \citeauthoryear {%
Womelsdorf%
, Anton-Erxleben%
, Pieper%
\BCBL {}\ \BBA {} Treue%
}{%
Womelsdorf%
\ \protect \BOthers {.}}{%
{\protect \APACyear {2006}}%
}]{%
womelsdorf2006dynamic}
\APACinsertmetastar {%
womelsdorf2006dynamic}%
\begin{APACrefauthors}%
Womelsdorf, T.%
, Anton-Erxleben, K.%
, Pieper, F.%
\BCBL {}\ \BBA {} Treue, S.%
\end{APACrefauthors}%
\unskip\
\newblock
\APACrefYearMonthDay{2006}{}{}.
\newblock
{\BBOQ}\APACrefatitle {Dynamic shifts of visual receptive fields in cortical area MT by spatial attention} {Dynamic shifts of visual receptive fields in cortical area mt by spatial attention}.{\BBCQ}
\newblock
\APACjournalVolNumPages{Nature neuroscience}{9}{9}{1156--1160}.
\PrintBackRefs{\CurrentBib}

\bibitem [\protect \citeauthoryear {%
Yamins%
\ \protect \BOthers {.}}{%
Yamins%
\ \protect \BOthers {.}}{%
{\protect \APACyear {2014}}%
}]{%
yamins2014performance}
\APACinsertmetastar {%
yamins2014performance}%
\begin{APACrefauthors}%
Yamins, D\BPBI L.%
, Hong, H.%
, Cadieu, C\BPBI F.%
, Solomon, E\BPBI A.%
, Seibert, D.%
\BCBL {}\ \BBA {} DiCarlo, J\BPBI J.%
\end{APACrefauthors}%
\unskip\
\newblock
\APACrefYearMonthDay{2014}{}{}.
\newblock
{\BBOQ}\APACrefatitle {Performance-optimized hierarchical models predict neural responses in higher visual cortex} {Performance-optimized hierarchical models predict neural responses in higher visual cortex}.{\BBCQ}
\newblock
\APACjournalVolNumPages{Proceedings of the national academy of sciences}{111}{23}{8619--8624}.
\PrintBackRefs{\CurrentBib}

\bibitem [\protect \citeauthoryear {%
Yang%
\ \protect \BOthers {.}}{%
Yang%
\ \protect \BOthers {.}}{%
{\protect \APACyear {2023}}%
}]{%
yang2023unimo}
\APACinsertmetastar {%
yang2023unimo}%
\begin{APACrefauthors}%
Yang, H.%
, Gao, C.%
, L{\'\i}u, H.%
, Xiao, X.%
, Zhao, Y.%
\BCBL {}\ \BBA {} Qin, B.%
\end{APACrefauthors}%
\unskip\
\newblock
\APACrefYearMonthDay{2023}{}{}.
\newblock
{\BBOQ}\APACrefatitle {UNIMO-3: Multi-granularity Interaction for Vision-Language Representation Learning} {Unimo-3: Multi-granularity interaction for vision-language representation learning}.{\BBCQ}
\newblock
\APACjournalVolNumPages{arXiv preprint arXiv:2305.13697}{}{}{}.
\PrintBackRefs{\CurrentBib}

\bibitem [\protect \citeauthoryear {%
Yue%
, Robert%
\BCBL {}\ \BBA {} Ungerleider%
}{%
Yue%
\ \protect \BOthers {.}}{%
{\protect \APACyear {2020}}%
}]{%
yue2020curvature}
\APACinsertmetastar {%
yue2020curvature}%
\begin{APACrefauthors}%
Yue, X.%
, Robert, S.%
\BCBL {}\ \BBA {} Ungerleider, L\BPBI G.%
\end{APACrefauthors}%
\unskip\
\newblock
\APACrefYearMonthDay{2020}{}{}.
\newblock
{\BBOQ}\APACrefatitle {Curvature processing in human visual cortical areas} {Curvature processing in human visual cortical areas}.{\BBCQ}
\newblock
\APACjournalVolNumPages{NeuroImage}{222}{}{117295}.
\PrintBackRefs{\CurrentBib}

\bibitem [\protect \citeauthoryear {%
Zeki%
}{%
Zeki%
}{%
{\protect \APACyear {1973}}%
}]{%
zeki1973colour}
\APACinsertmetastar {%
zeki1973colour}%
\begin{APACrefauthors}%
Zeki, S\BPBI M.%
\end{APACrefauthors}%
\unskip\
\newblock
\APACrefYearMonthDay{1973}{}{}.
\newblock
{\BBOQ}\APACrefatitle {Colour coding in rhesus monkey prestriate cortex} {Colour coding in rhesus monkey prestriate cortex}.{\BBCQ}
\newblock
\APACjournalVolNumPages{Brain research}{53}{2}{422--427}.
\PrintBackRefs{\CurrentBib}

\end{thebibliography}

\appendix
\renewcommand{\thefigure}{A\arabic{figure}}
\renewcommand{\thetable}{A\arabic{table}}
\setcounter{figure}{0}  
\setcounter{table}{0}
\newpage
\section{Appendix}

\begin{table*}[!ht]
\begin{center} 
\caption{Performance (Test Accuracies as Normalized Pearson Correlation) of various Task Optimized vision models with Linear Ridge (R), Spatial-Feature Factorized Linear (F), Semantic Spatial Transformer (S) and Gaussian2D (G) readouts} 
\label{tab:Task_Optimized_model_comp} 
\vskip 0.12in
\begin{tabular}{|c|c|c|c|c|c|c|c|c|c|c|c|}
\hline
\multicolumn{2}{|c|}{\textbf{Model Details}} & \multicolumn{10}{c|}{\textbf{Visual Cortex Region}} \\ \hline
    \textbf{Layers} & \textbf{Readout} & \textbf{V1v} & \textbf{V1d} & \textbf{V2v} & \textbf{V2d} & \textbf{V3v} & \textbf{V3d} & \textbf{V4} & \textbf{Ventral} & \textbf{Dorsal} & \textbf{Lateral} \\ \hline
    \multicolumn{12}{|c|}{\textbf{ResNet 50}} \\ \hline
    \multirow{4}{*}{1} & R & 0.6009 & 0.5695 & 0.5168 & 0.4783 & 0.4612 & 0.4543 & 0.4085 & 0.2958 & 0.3101 & 0.2508 \\ 
        & G & 0.5935 & 0.5634 & 0.5110 & 0.4238 & 0.4135 & 0.4148 & 0.3758 & 0.2236 & 0.1928 & 0.1940 \\ 
        & F & 0.8041 & 0.7627 & 0.7321 & 0.6950 & 0.6517 & 0.6540 & 0.5771 & 0.3252 & 0.3318 & 0.2709 \\ 
        & S (Ours) & \underline{0.8498} & \underline{0.8022} & \underline{0.7860} & \underline{0.7501} & \underline{0.7559} & \underline{0.7461} & \underline{0.7410} & \underline{0.5763} & \underline{0.6208} & \underline{0.5652} \\ \hline
    \multirow{4}{*}{2} & R & 0.5618 & 0.6535 & 0.6276 & 0.4677 & 0.4564 & 0.4485 & 0.4157 & 0.3628 & 0.3796 & 0.3131 \\ 
        & G & 0.6478 & 0.5827 & 0.5694 & 0.5086 & 0.4975 & 0.4861 & 0.4898 & 0.2958 & 0.2797 & 0.2593 \\ 
        & F & 0.8142 & 0.7728 & 0.7601 & 0.7302 & 0.6956 & 0.7110 & 0.6403 & 0.4034 & 0.4116 & 0.3515 \\ 
        & S (Ours) & \underline{\textbf{0.8507}} & \underline{\textbf{0.8083}} & \underline{\textbf{0.8057}} & \underline{\textbf{0.7603}} & \underline{\textbf{0.7612}} & \underline{\textbf{0.7763}} & \underline{0.7601} & \underline{0.5813} & \underline{0.6241} & \underline{0.5667} \\ \hline
    \multirow{4}{*}{3} & R & 0.6599 & 0.6413 & 0.6426 & 0.6014 & 0.6051 & 0.6237 & 0.6138 & 0.5022 & 0.5657 & 0.4689 \\ 
        & G & 0.6607 & 0.6110 & 0.6359 & 0.5920 & 0.6270 & 0.6205 & 0.6526 & 0.4991 & 0.5296 & 0.4671 \\ 
        & F & \underline{0.8046} & \underline{0.7666} & \underline{0.7705} & \underline{0.7482} & 0.7465 & 0.7675 & 0.7540 & 0.5751 & 0.6277 & 0.5384 \\ 
        & S (Ours) & 0.7898 & 0.7393 & 0.7643 & 0.7193 & \underline{0.7496} & \underline{0.7495} & \underline{\textbf{0.7674}} & \underline{\textbf{0.6105}} & \underline{\textbf{0.6606}} & \underline{\textbf{0.5823}} \\ \hline
    \multirow{4}{*}{4 (all)} & R & 0.2812 & 0.2577 & 0.2583 & 0.2556 & 0.2880 & 0.2433 & 0.3132 & 0.3006 & 0.2922 & 0.2820 \\ 
        & G & 0.5170 & 0.4671 & 0.4810 & 0.4318 & 0.4821 & 0.4787 & 0.5442 & 0.4764 & 0.4702 & 0.4704 \\ 
        & F & 0.5922 & 0.5488 & 0.5606 & 0.5297 & 0.5542 & 0.5659 & 0.5612 & 0.4525 & 0.4741 & 0.4269 \\ 
        & S (Ours) & \underline{0.6989} & \underline{0.6504} & \underline{0.6746} & \underline{0.6487} & \underline{0.6743} & \underline{0.6791} & \underline{0.6814} & \underline{0.5857} & \underline{0.6337} & \underline{0.5809} \\ \hline
    \multicolumn{12}{|c|}{\textbf{AlexNet}} \\ \hline
    \multirow{4}{*}{1} & R & 0.6359 & 0.6320 & 0.5844 & 0.5403 & 0.5268 & 0.5178 & 0.4795 & 0.3134 & 0.3275 & 0.2727 \\ 
        & G & 0.6520 & 0.6009 & 0.5539 & 0.5197 & 0.4649 & 0.4489 & 0.4550 & 0.3156 & 0.3054 & 0.2662 \\ 
        & F & 0.7253 & 0.6897 & 0.6479 & 0.6136 & 0.5678 & 0.5841 & 0.5300 & 0.3170 & 0.3178 & 0.2763 \\ 
        & S (Ours) & \underline{0.7590} & \underline{0.7159} & \underline{0.7229} & \underline{0.6662} & \underline{0.6934} & \underline{0.6764} & \underline{0.7004} & \underline{0.5594} & \underline{0.6072} & \underline{0.5556} \\ \hline
    \multirow{4}{*}{2} & R & 0.5822 & 0.5550 & 0.5268 & 0.4951 & 0.4919 & 0.4855 & 0.4715 & 0.2924 & 0.2949 & 0.2485 \\ 
        & G & 0.6459 & 0.6221 & 0.5883 & 0.5489 & 0.5439 & 0.5357 & 0.5278 & 0.3688 & 0.3399 & 0.3271 \\ 
        & F & 0.7325 & 0.6923 & 0.6704 & 0.6396 & 0.6168 & 0.6287 & 0.5876 & 0.3864 & 0.3822 & 0.3322 \\ 
        & S (Ours) & \underline{0.7710} & \underline{\textbf{0.7288}} & \underline{0.7273} & \underline{0.6950} & \underline{0.7043} & \underline{\textbf{0.7117}} & \underline{\textbf{0.7169}} & \underline{0.5705} & \underline{0.6002} & \underline{0.5408} \\ \hline
    \multirow{4}{*}{3} & R & 0.5951 & 0.5722 & 0.5554 & 0.5260 & 0.5234 & 0.5313 & 0.5197 & 0.3389 & 0.3392 & 0.2879 \\ 
        & G & 0.6311 & 0.6150 & 0.5997 & 0.5705 & 0.5597 & 0.5629 & 0.5719 & 0.4289 & 0.4072 & 0.3888 \\
        & F & 0.7419 & 0.7055 & 0.7033 & 0.6713 & 0.6655 & 0.6864 & 0.6594 & 0.4618 & 0.4711 & 0.4098 \\ 
        & S (Ours) & \underline{\textbf{0.7634}} & \underline{0.7236} & \underline{\textbf{0.7327}} & \underline{\textbf{0.6961}} & \underline{\textbf{0.7065}} & \underline{0.7092} & \underline{0.7148} & \underline{0.5694} & \underline{0.6071} & \underline{0.5326} \\ \hline
    \multirow{4}{*}{4} & R & 0.6145 & 0.5830 & 0.5834 & 0.5527 & 0.5733 & 0.5677 & 0.5632 & 0.4123 & 0.4181 & 0.3486 \\ 
        & G & 0.6357 & 0.6038 & 0.5994 & 0.5677 & 0.5684 & 0.5795 & 0.5908 & 0.4601 & 0.4827 & 0.4220 \\ 
        & F & 0.7325 & 0.6933 & 0.6989 & 0.6724 & 0.6735 & 0.6895 & 0.6758 & 0.5066 & 0.5323 & 0.4559 \\
        & S (Ours) & \underline{0.7444} & \underline{0.7070} & \underline{0.7173} & \underline{0.6816} & \underline{0.7037} & \underline{0.7108} & \underline{0.7129} & \underline{0.5688} & \underline{0.6214} & \underline{0.5458} \\ \hline
    \multirow{4}{*}{5 (all)} & R & 0.4931 & 0.4798 & 0.4703 & 0.4474 & 0.4560 & 0.4609 & 0.4662 & 0.3717 & 0.3806 & 0.3394 \\
        & G & 0.5605 & 0.5652 & 0.5136 & 0.5193 & 0.4946 & 0.5231 & 0.5260 & 0.4523 & 0.4334 & 0.4201 \\ 
        & F & 0.6889 & 0.6339 & 0.6679 & 0.6183 & 0.6602 & 0.6502 & 0.6833 & 0.5803 & 0.6347 & 0.5797 \\ 
        & S (Ours) & \underline{0.7168} & \underline{0.6653} & \underline{0.6859} & \underline{0.6481} & \underline{0.6855} & \underline{0.6797} & \underline{0.7156} & \underline{\textbf{0.6003}} & \underline{\textbf{0.6443}} & \underline{\textbf{0.5965}} \\
\hline
\multicolumn{12}{|c|}{\textbf{ConvNext Base}} \\ \hline
\multirow{1}{*}{1} & S & \textbf{0.8238} & 0.7744 & 0.775 & 0.7324 & 0.7334 & 0.7361 & 0.7257 & 0.5579 & 0.5948 & 0.5186 \\  \hline
\multirow{1}{*}{2} & S & 0.8194 & \textbf{0.7762} & \textbf{0.7753} & \textbf{0.7425} & \textbf{0.7520} & \textbf{0.7595} & \textbf{0.7536} & 0.5745 & 0.6172 & 0.5521 \\  \hline
\multirow{1}{*}{3} & S & 0.6697 & 0.6345 & 0.6445 & 0.6057 & 0.6504 & 0.6600 & 0.6858 & \textbf{0.5756} & 0.6392 & 0.5836 \\  \hline
\multirow{1}{*}{4 (all)} & S & 0.6688 & 0.6209 & 0.6394 & 0.5914 & 0.6394 & 0.6351 & 0.6742 & 0.5761 & \textbf{0.6374} & \textbf{0.5679} \\  \hline
\multicolumn{12}{|c|}{\textbf{Moco V2}} \\ \hline
\multirow{1}{*}{1} & S & 0.8379 & 0.7735 & 0.7898 & 0.7370 & 0.7505 & 0.7459 & 0.7552 & 0.5828 & 0.6083 & 0.5635 \\  \hline
\multirow{1}{*}{2} & S & \textbf{0.8405} & \textbf{0.7967} & \textbf{0.8018} & \textbf{0.7630} & \textbf{0.7632} & \textbf{0.7736} & 0.7589 & 0.5917 & 0.6366 & 0.5758 \\  \hline
\multirow{1}{*}{3} & S & 0.8066 & 0.7604 & 0.7791 & 0.7340 & 0.7589 & 0.7722 & \textbf{0.7649} & 0.6082 & \textbf{0.6621} & \textbf{0.5850} \\  \hline
\multirow{1}{*}{4 (all)} & S & 0.7111 & 0.6586 & 0.6793 & 0.6457 & 0.6735 & 0.6790 & 0.6980 & \textbf{0.6038} & 0.6641 & 0.6019 \\  \hline
\end{tabular} 
\end{center} 
\end{table*}

\begin{table*}[!ht]
\begin{center} 
\caption{Performance (Test Accuracies as Normalized Pearson Correlation) of Response Optimized vision models with Linear Ridge (R), Spatial-Feature Factorized Linear (F), Semantic Spatial Transformer (S) and Gaussian2D (G) readouts} 
\label{tab:Response_Optimized_Vision_model_comp} 
\vskip 0.12in
\begin{tabular}{|c|c|c|c|c|c|c|c|c|c|c|}
\hline
\textbf{Readout} & \textbf{V1v} & \textbf{V1d} & \textbf{V2v} & \textbf{V2d} & \textbf{V3v} & \textbf{V3d} & \textbf{V4} & \textbf{Ventral} & \textbf{Dorsal} & \textbf{Lateral} \\
\hline
R & 0.7746 & 0.7427 & 0.7299 & 0.6906 & 0.6867 & 0.6865 & 0.6551 & 0.4657 & 0.4824 & 0.4372 \\ 
G & 0.7306 & 0.6744 & 0.6746 & 0.6253 & 0.6326 & 0.6104 & 0.6297 & 0.4784 & 0.4728 & 0.4545 \\
F & 0.83154 & 0.7926 & 0.7795 & 0.7419 & 0.7268 & 0.7323 & 0.7085 & 0.4847 & 0.4831 & 0.4504 \\ 
S (Ours) & \textbf{0.8698} & \textbf{0.8340} & \textbf{0.8302} & \textbf{0.7919} & \textbf{0.7808} & \textbf{0.7913} & \textbf{0.7729} & \textbf{0.5796} & \textbf{0.6089} & \textbf{0.5638} \\ \hline
\hline
\end{tabular} 
\end{center} 
\end{table*}

\begin{table*}[!ht]
\begin{center} 
\caption{Performance (Test Accuracies as Normalized Pearson Correlation) of language models (C: CLIP, M: MPNET, G-XL: GPT2-XL) with Linear Ridge (R), Spatial-Feature Factorized Linear (F), Semantic Spatial Transformer (S) and Gaussian2D (G) readouts} 
\label{tab:Response_Optimized_lANGUAGE_model_comp} 
\vskip 0.12in
\begin{tabular}{|c|c|c|c|c|c|c|c|c|c|c|c|}
\hline
\multicolumn{2}{|c|}{\textbf{Model Details}} & \multicolumn{10}{c|}{\textbf{Visual Cortex Region}} \\ \hline
    \textbf{LLM} & \textbf{Readout} & \textbf{V1v} & \textbf{V1d} & \textbf{V2v} & \textbf{V2d} & \textbf{V3v} & \textbf{V3d} & \textbf{V4} & \textbf{Ventral} & \textbf{Dorsal} & \textbf{Lateral} \\ \hline
    \multicolumn{12}{|c|}{\textbf{Single Caption Models}} \\
\hline
C & R & \textbf{0.3974} & \textbf{0.3779} & \textbf{0.3809} & \textbf{0.3702} & \textbf{0.4093} & \textbf{0.4119} & \textbf{0.4882} & 0.5661 & 0.6243 & 0.5920 \\ 
M & R & 0.3931 & 0.3738 & 0.3738 & 0.3687 & 0.4031 & 0.4077 & 0.4873 &  \textbf{0.5672} & \textbf{0.6269} & \textbf{0.6126} \\ 
G-XL & R & 0.3791 & 0.3642 & 0.3653 & 0.3540 & 0.3953 & 0.4036 & 0.4773 & 0.5638 & 0.6162 & 0.6007 \\ \hline
\multicolumn{12}{|c|}{\textbf{Dense Caption Models}} \\ \hline
\multirow{3}{*}{C} & R & 0.6597 & 0.6154 & 0.6551 & 0.5953 & 0.6371 & 0.6322 & 0.6621 & 0.5807 & 0.6201 & 0.5761 \\ 
    & G & 0.6783 & 0.6277 & 0.6682 & 0.6207 & 0.6644 & 0.6531 & 0.6905 & 0.5980 & 0.6491 & 0.5943 \\ 
    & F & 0.6919 & 0.6329 & 0.6721 & 0.6183 & 0.6603 & 0.6572 & 0.6927 & 0.5915 & 0.6365 & 0.5781 \\ 
    & S (Ours) & \underline{0.7196} & \underline{0.6590} & \underline{0.6903} & \underline{0.6457} & \underline{0.6897} & \underline{0.6774} & \underline{0.7167} & \underline{0.5953} & \underline{\textbf{0.6562}} & \underline{0.6001} \\ \hline
\multirow{3}{*}{M} & R & 0.6557 & 0.5941 & 0.6325 & 0.5732 & 0.6162 &  0.6207 & 0.6493 & 0.5679 & 0.5831 & 0.5502 \\ 
    & G & 0.6840 & 0.6261 & 0.6659 & 0.6207 & 0.6583 & 0.6519 & 0.6928 & 0.5934 & 0.6441 & 0.5894 \\ 
    & F & 0.6889 & 0.6339 & 0.6679 & 0.6183 & 0.6602 & 0.6502 & 0.6833 & 0.5803 & 0.6347 & 0.5797 \\ 
    & S (Ours) & \underline{0.7168} & \underline{0.6653} & \underline{0.6859} & \underline{0.6481} & \underline{0.6855} & \underline{0.6797} & \underline{0.7156} & \underline{\textbf{0.6003}} & \underline{0.6443} & \underline{0.5965} \\ \hline
\multirow{3}{*}{G-XL} & R & 0.6738 & 0.6272 & 0.6586 & 0.6136 & 0.6625 &  0.6504 & 0.6862 & 0.5881 & 0.6380 & 0.5732 \\ 
    & G & 0.6895 & 0.6284 & 0.6717 & 0.6203 & 0.6605 & 0.6631 & 0.6980 & 0.5941 & \underline{0.6501} & 0.6003 \\ 
    & F & 0.6940 & 0.6386 & 0.6716 & 0.6275 & 0.6597 & 0.6636 & 0.6974 & 0.5874 & 0.6381 & 0.5832 \\ 
    & S (Ours) & \underline{\textbf{0.7253}} & \underline{\textbf{0.6653}} & \underline{\textbf{0.7038}} & \underline{\textbf{0.6619}} & \underline{\textbf{0.6956}} & \underline{\textbf{0.6939}} & \underline{\textbf{0.7242}} & \underline{0.5974} & 0.6487 & \underline{\textbf{0.6023}} \\ \hline
\end{tabular} 
\end{center} 
\end{table*}


\begin{figure*}[h!]
\centering
\includegraphics[width=\linewidth]{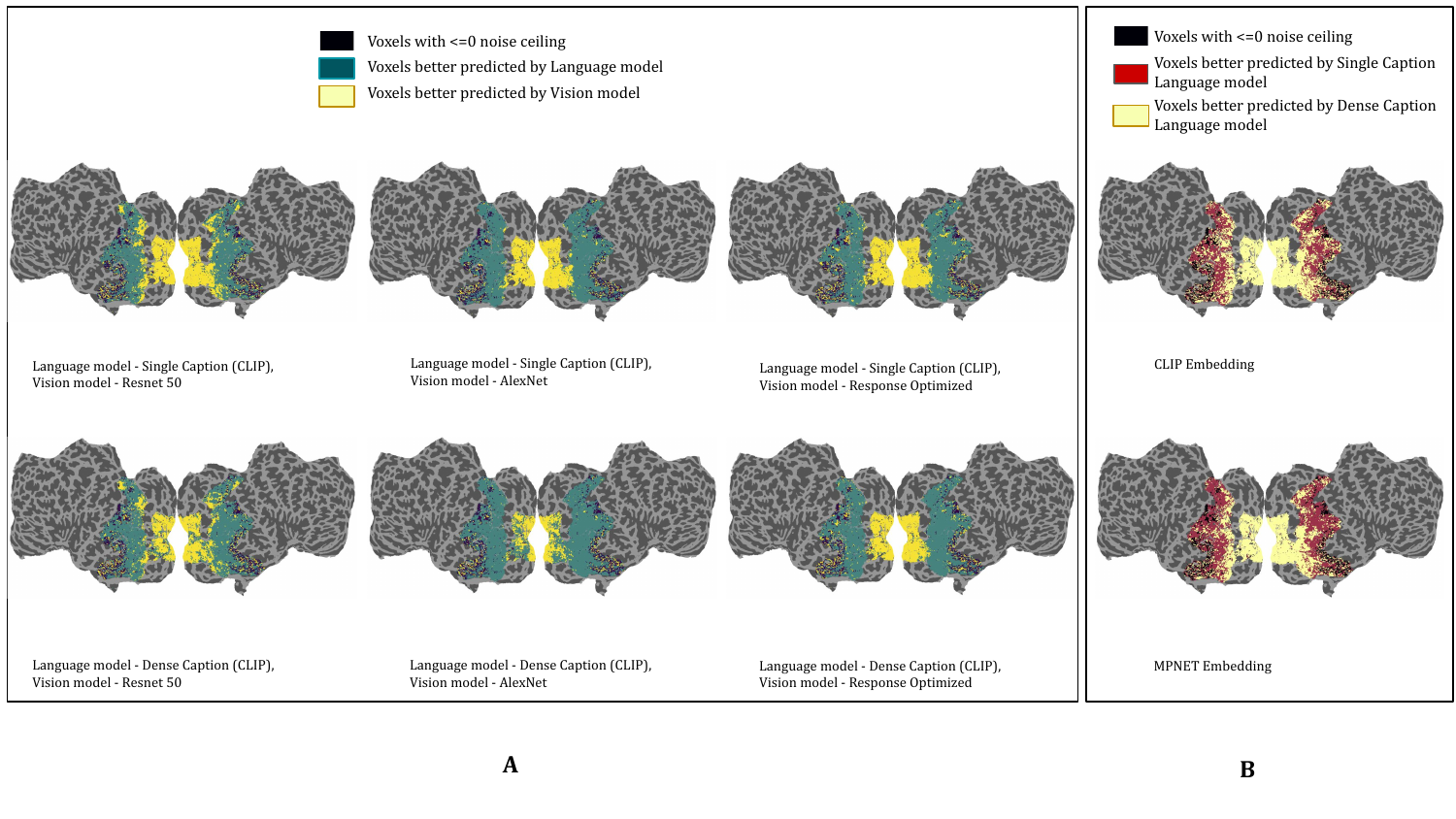}
\caption{A - Brain Visualizations showing voxels that are better predicted by vision and language models, all using Ridge Linear readouts, B - Brain Visualizations
showing voxels that are better predicted by single caption and dense caption language models, all using Ridge Linear readouts}
\label{fig:vision_lang_ridge_linear}
\end{figure*}

\begin{figure*}[h!]
\centering
\includegraphics[width=\linewidth]{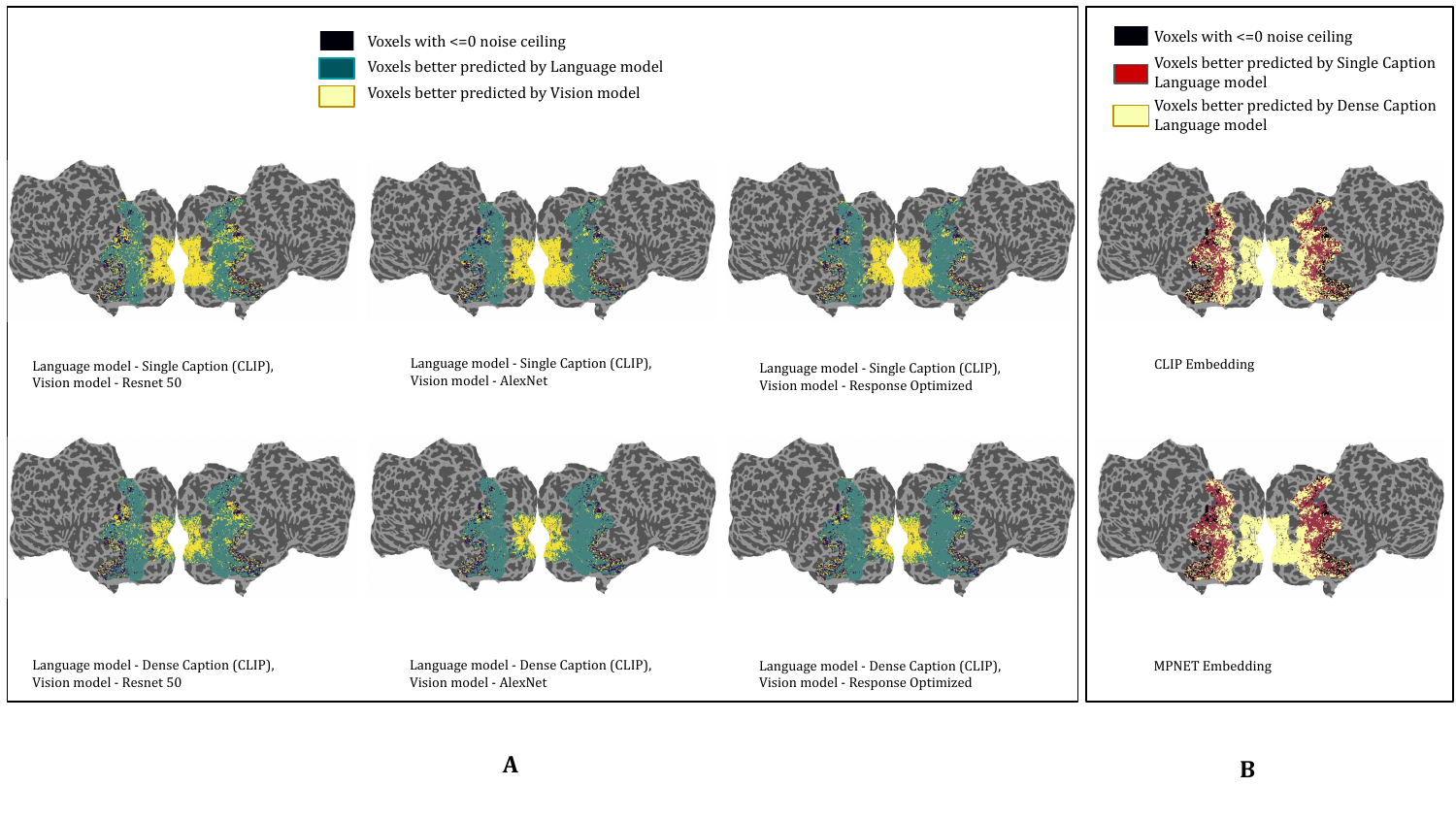}
\caption{A - Brain Visualizations showing voxels that are better predicted by vision and language models, all using Gaussian2D readouts, B - Brain Visualizations
showing voxels that are better predicted by single caption and dense caption language models, all using gaussian2D readouts}
\label{fig:vision_lang_gaussian2D}
\end{figure*}

\begin{figure*}[h!]
\centering
\includegraphics[width=\linewidth]{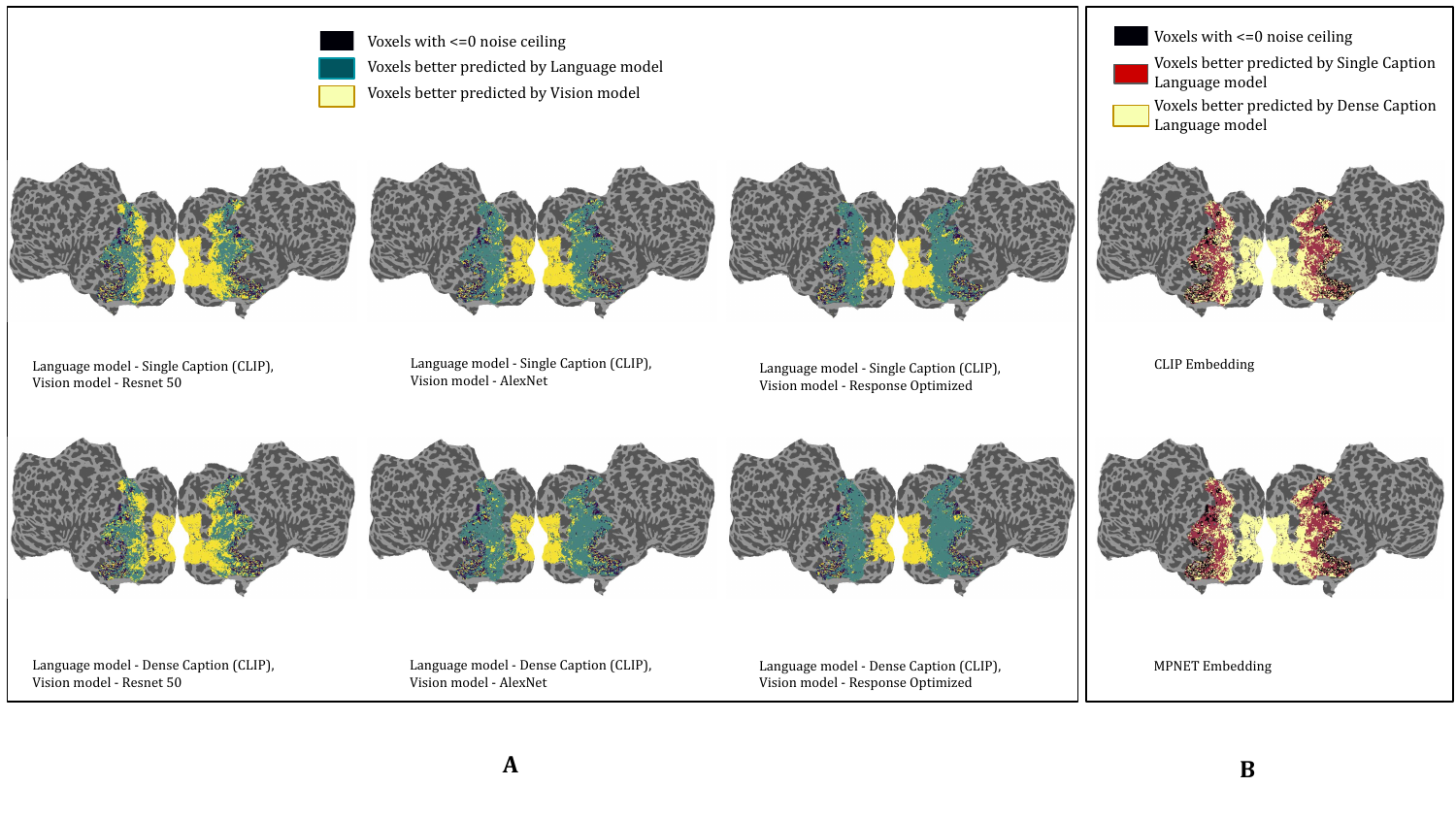}
\caption{A - Brain Visualizations showing voxels that are better predicted by Vision and language models, all using Spatial-Feature Factorized Linear readouts, B - Brain Visualizations
showing voxels that are better predicted by single caption and dense caption language models, all using Spatial-Feature Factorized Linear readouts}
\label{fig:vision_lang_spatial_linear}
\end{figure*}

\subsection{Natural Scenes Dataset}
\label{subsec:nsd}
A detailed description of the Natural Scenes Dataset (NSD; http://naturalscenesdataset.org) is provided elsewhere \citep{allen2022massive}. The NSD dataset contains measurements of fMRI responses from 8 participants who each viewed 9,000–10,000 distinct color natural scenes (22,000–30,000 trials) over the course of 30–40 scan sessions. Scanning was conducted at 7T using whole-brain gradient-echo EPI at 1.8-mm resolution and 1.6-s repetition time. Images were taken from the Microsoft Common Objects in Context (COCO) database \citep{lin2014microsoft}, square cropped, and presented at a size of 8.4° x 8.4°. A special set of 1,000 images were shared across subjects; the remaining images were mutually exclusive across subjects. Images were presented for 3 s with 1-s gaps in between images. Subjects fixated centrally and performed a long-term continuous recognition task on the images. The fMRI data were pre-processed by performing one temporal interpolation (to correct for slice time differences) and one spatial interpolation (to correct for head motion). A general linear model was then used to estimate single-trial beta weights. Cortical surface reconstructions were generated using FreeSurfer, and both volume- and surface-based versions of the beta weights were created. In this study, we analyze manually defined regions of interest (ROIs) across both early and higher-level visual cortical areas. For early visual areas, we focus on ROIs delineated based on the results of the population receptive field (pRF) experiment - V1v, V1d, V2v, V2d, V3v, V3d, and hV4. For higher level visual cortex regions, we target the ventral, dorsal, and lateral streams, as defined by the streams atlas.

\textbf{Noise Ceiling Estimation in NSD} - Noise ceiling for every voxel represents the performance of the ``true" model underlying the generation of the responses (the best achievable accuracy) given the noise in the fMRI measurements. They were computed using the standard procedure followed in ~\citep{allen2022massive} by considering the variability in voxel responses across repeat scans. The dataset contains 3 different responses to each stimulus image for every voxel. In the estimation framework, the variance of the responses, $\sigma_{\text{response}}^2$, are split into two components, the measurement noise $\sigma_{\text{noise}}^2$ and the variability between images of the noise free responses $\sigma_{\text{signal}}^2$.
\begin{align*}
    \hat{\sigma}^2_{\text{response}} = \hat{\sigma}^2_{\text{signal}}  + \hat{\sigma}^2_{\text{noise}}
\end{align*}
An estimate of the variability of the noise is given as $\hat{\sigma}^2_{\text{noise}}  = \frac{1}{n}\sum_{i=1}^n\text{Var}(\beta_i)$, where i denotes the image (among $n$ images) and $\text{Var}(\beta_i)$ denotes the variance of the response across repetitions of the same image. An estimate of the variability of the noise free signal is then given as,
\begin{align*}
    \hat{\sigma}^2_{\text{signal}} = \hat{\sigma}^2_{\text{response}}  - \hat{\sigma}^2_{\text{noise}}
\end{align*}
Since the measured responses were z-scored, $\hat{\sigma}^2_{\text{response}}=1$ and $\hat{\sigma}^2_{\text{signal}} = 1 - \hat{\sigma}^2_{\text{noise}} $. The noise ceiling (n.c.) expressed in correlation units is thus given as $n.c. = \sqrt{\frac{\hat{\sigma}^2_{\text{signal}}}{\hat{\sigma}^2_{\text{signal}} + \hat{\sigma}^2_{\text{noise}}}}$. The models were evaluated in terms of their ability to explain the average response across 3 trials (i.e., repetitions) of the stimulus. To account for this trial averaging, the noise ceiling is expressed as $n.c. = \sqrt{\frac{\hat{\sigma}^2_{\text{signal}}}{\hat{\sigma}^2_{\text{signal}} + \hat{\sigma}^2_{\text{noise}}/3}}$. We computed noise ceiling using this formulation for every voxel in each subject and expressed the noise-normalized prediction accuracy (R) as a fraction of this noise ceiling. 

\begin{figure*}[h!]
\centering
\includegraphics[width=\linewidth, trim={0 0 0 0}, clip]{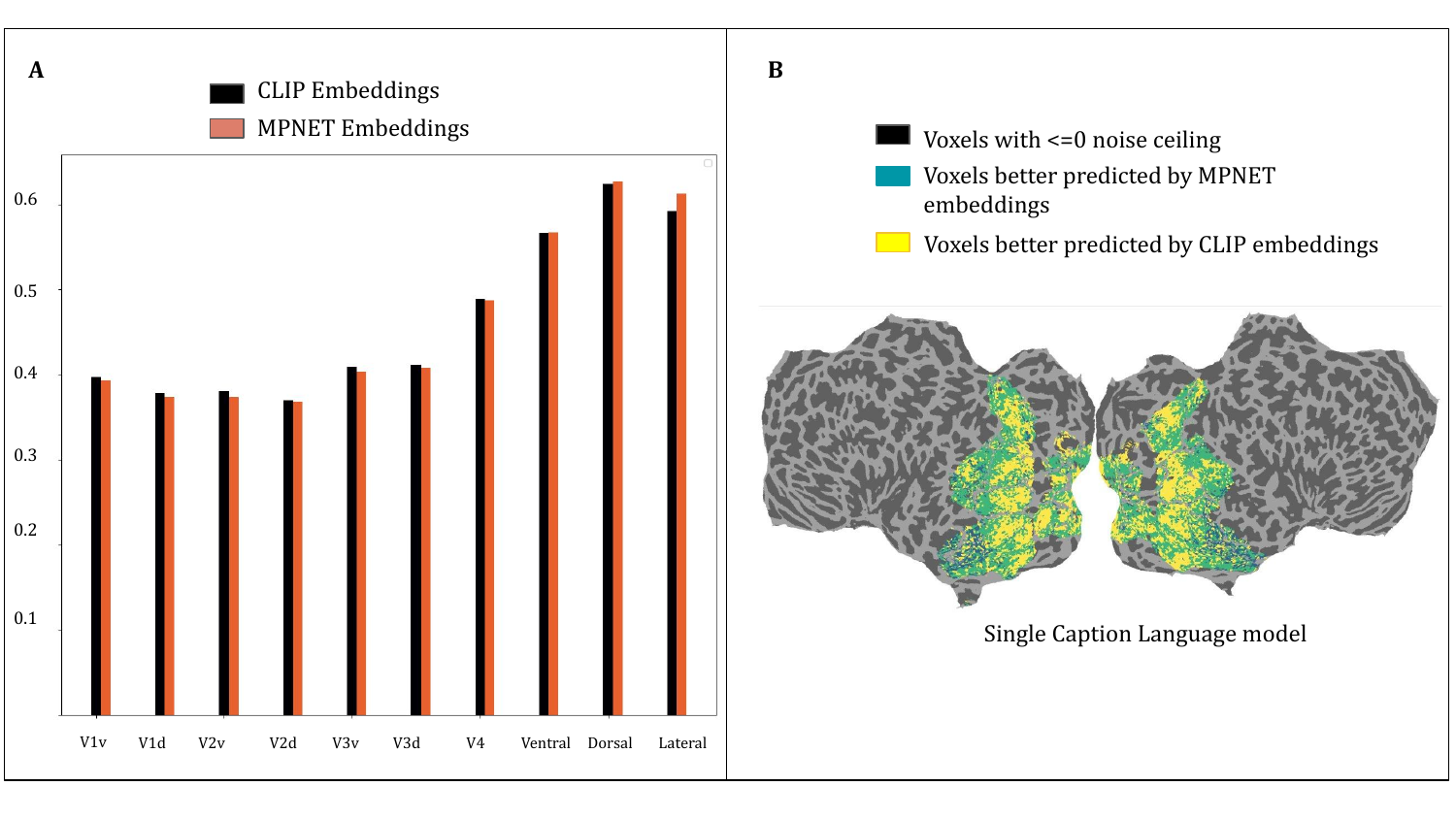}
\caption{Comparison of Unimodal and Multimodal embeddings in Language models, A - Test Accuracy (Normalized Pearson Correlation) on held out dataset using Single Caption Language encoders with CLIP and MPNET embeddings, B - Brain Visualization showing regions better predicted by each encoder in Single Caption Language models}
\label{fig:clip_mpnet}
\end{figure*}

\subsection{Unimodal versus multimodal embeddings in language models}
\label{subsec:cm}

As outlined in the previous section, the higher-level regions of the ventral, dorsal, and lateral visual streams exhibit heightened sensitivity to broad semantic information that captures the overall meaning of a scene, as opposed to specific visual details or a combination of visual and spatial features. These regions are best modeled by single-caption language models. To investigate this further, we examine the performance of models using unimodal encoders like MPNET, which are trained exclusively on language, and multimodal encoders like CLIP, trained on both language and visual data.  In the higher regions of the ventral, dorsal, and lateral streams, models using MPNET encoders slightly outperform those with CLIP encoders by 0.5\%.  This marginal advantage in the higher regions may be attributed to MPNET's optimization for capturing rich semantic nuances from text, aligning well with the language-sensitive nature of these brain regions. On the other hand, in the lower visual regions, where responses are more strongly driven by visual inputs, CLIP encoders hold a small advantage of 1\% over MPNET, likely due to their integration of visual knowledge. However, this trend does not hold in dense caption language models, where the performance of both encoders is comparable.

\begin{figure*}[h!]
\centering
\includegraphics[width=\linewidth, trim={0 130 0 0}, clip]{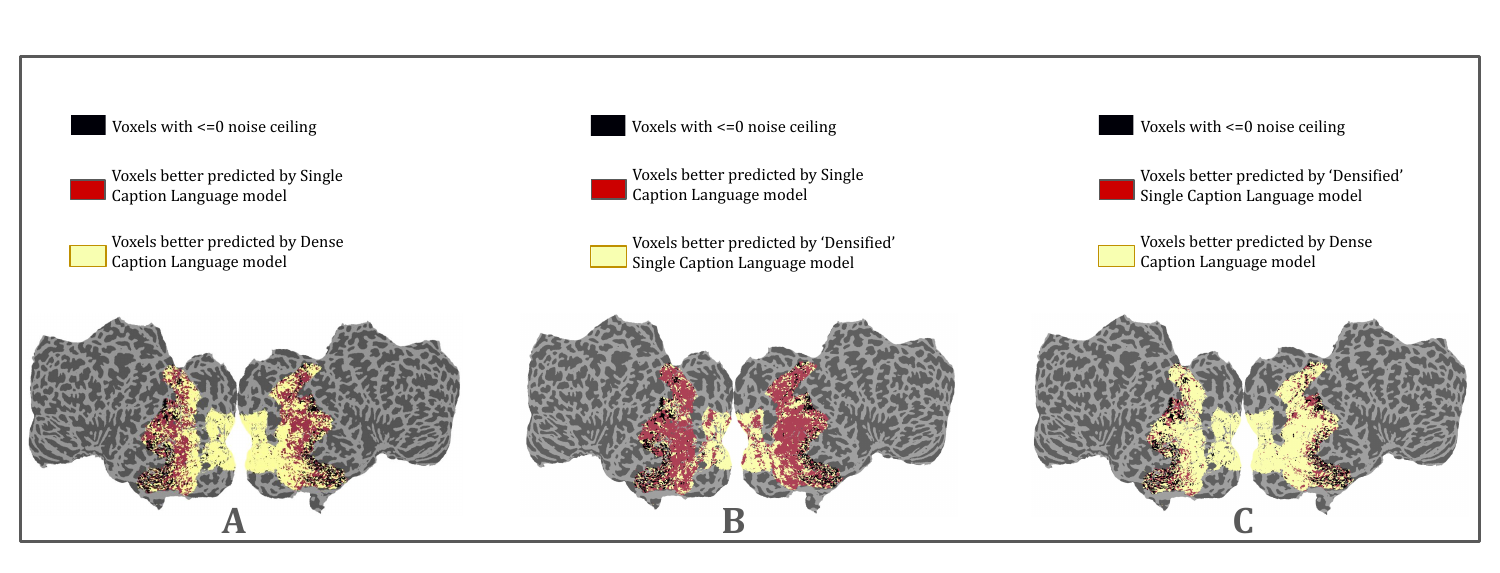}
\caption{A - Comparison of Single Caption Language models with Dense Caption Language models, B - Comparison of Single Caption Language models with 'Densified' Single Caption Language model, C - Comparison of 'Densified' Single Caption Language model with Dense Caption Language model}
\label{fig:densified_single_captions}
\end{figure*}

\subsection{The Necessity of Spatial Subdivision in Dense Captioning for Effective Visual Cortex Modeling}
\label{subsec:dense_captions}
 We further investigated whether the observed differences between dense and global captioning are due to (a) the spatial subdivision of the image (Hypothesis 1) or the increased semantic detail in dense captions (Hypothesis 2). The original idea behind using dense captions was to provide spatial information in addition to semantic information in the form of captions, and subdividing the image into equal sized grids and getting captions for each grid was one of the easiest and most intuitive ways to do that. 

We further tried generating more comprehensive single captions of the image using existing LLMs, however none of them were able to provide more information than those already present in the original MS-COCO dataset. In an attempt to densify the single captions, we thus adopted a different approach: for each image, we took the embeddings of dense captions generated for individual grid locations and averaged these embeddings to produce a single "aggregate dense caption" embedding.
 
On comparing single caption stimuli with ‘densified’ single caption stimuli (as opposed to the dense caption approach discussed in the paper) (Figure \ref{fig:densified_single_captions}), we saw a similar trend where the higher regions of the visual cortex were better modeled by single caption stimuli. However, the transition in sensitivity from dense to single caption in the middle regions of the ventral, dorsal and lateral stream that is so clearly pronounced when using dense captions is missing when using the above ‘densified’ single captions. Further comparing ‘densified’ single captions to dense captions (as proposed in the paper), we saw that the dense captions modeled the overall visual cortex better. Hence, we do feel that adding spatial information to the dense caption is necessary for building more accurate models, be it by sub-dividing the image into grids or via any other way.

\begin{table*}[!ht]
\begin{center} 
\caption{Performance (Test Accuracies as Normalized Pearson Correlation) of Spatial-Feature Factorized Linear Readout (F) with individual affine transformations applied to encoder feature maps (1) and spatial masks (2) separately, all with Response Optimized Vision models.} 
\label{tab:stn_sep_analysis} 
\vskip 0.12in
\begin{tabular}{|c|c|c|c|c|c|c|c|c|c|c|}
\hline
\textbf{Readout} & \textbf{V1v} & \textbf{V1d} & \textbf{V2v} & \textbf{V2d} & \textbf{V3v} & \textbf{V3d} & \textbf{V4} & \textbf{Ventral} & \textbf{Dorsal} & \textbf{Lateral} \\ \hline
      F & 0.83154 & 0.7926 & 0.7795 & 0.7419 & 0.7268 & 0.7323 & 0.7085 & 0.4847 & 0.4831 & 0.4504 \\ 
      F + 1 & 0.8596 & 0.8217 & 0.8179 & 0.7769 & 0.7705 & 0.7719 & 0.7659 & 0.5638 & 0.5962 & 0.5371 \\ 
      F + 2 & 0.8750 & 0.8409 & 0.8310 & 0.7948 & 0.7814 & 0.7958 & 0.7782 & 0.5865 & 0.6156 & 0.5641 \\ \hline
\end{tabular}
\end{center} 
\end{table*}

\begin{figure*}[h!]
\centering
\includegraphics[width=\linewidth, trim={0 130 0 0}, clip]{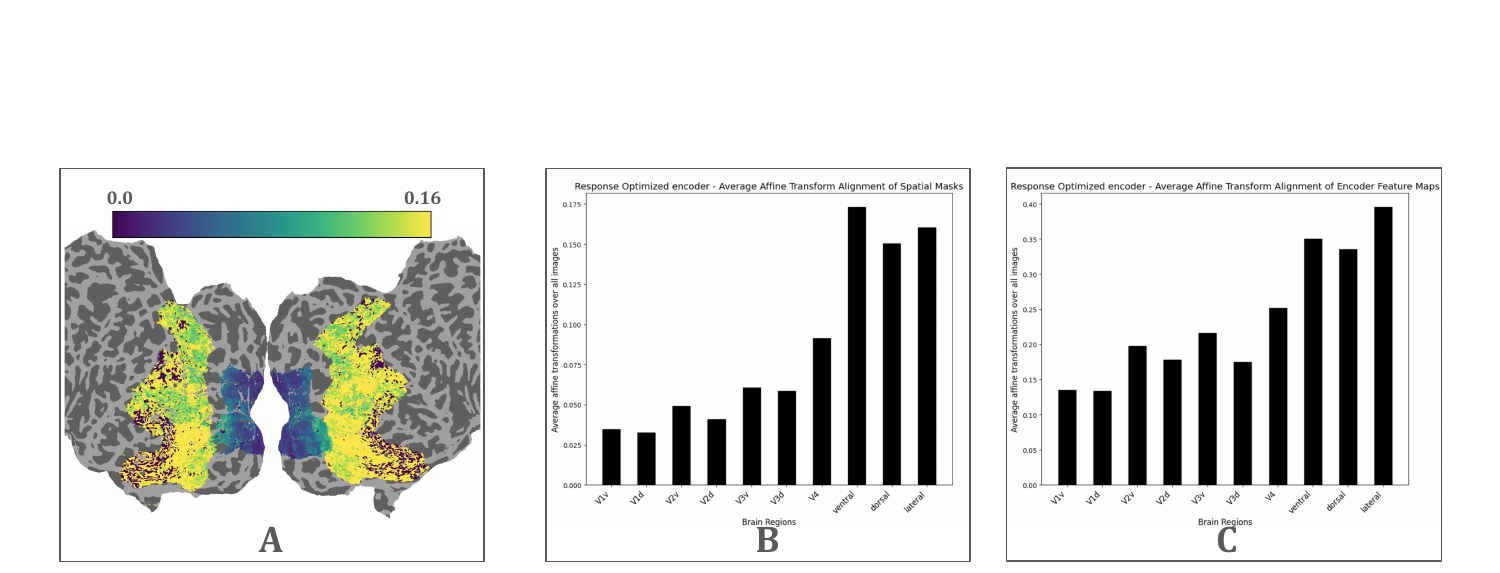}
\caption{A - Average spatial shifts of voxel spatial masks across all images, B - Mean spatial shifts for each brain region, comparing spatial masks across all images, C - Mean spatial shifts for each brain region, comparing feature maps across all images.}
\label{fig:affine_transform_alignment}
\end{figure*}

\subsection{Analyzing spatial modulation of Receptive Fields in visual cortex: Insights from STN Readouts}
\label{subsec:stn_readouts}

In an additional experiment focused on interpreting the STN readouts, we calculated the distance between the affine parameters corresponding to the spatial maps of each voxel for every image, relative to the mean affine parameters across all images (Figure \ref{fig:affine_transform_alignment}). The L2 norm of this vector was computed for each voxel. Across all encoders, we observed that stimulus-dependent spatial shifts of the receptive field increase from lower to higher visual regions. A similar trend emerged when calculating the average spatial shifts for each channel of the feature map across images for different regions. This trend further supports the idea that higher levels of the visual cortex benefit more from learned geometric invariances and exhibit greater spatial modulation of their visual receptive fields compared to lower visual cortex regions. This modulation includes phenomena such as receptive field expansion, contraction, or shifts in response to different stimuli.

\begin{table*}[!ht]
\begin{center} 
\caption{Performance (Analysis of the effect of channel size on the improvement introduced by Semantic Spatial Transformer Readout (S) over Spatial-Linear Factorized Readouts (F), all with Response Optimized Vision models} 
\label{tab:stn_channel_analysis} 
\vskip 0.12in
\begin{tabular}{|c|c|c|c|c|c|c|c|c|c|c|}
\hline
\textbf{Readout} & \textbf{V1v} & \textbf{V1d} & \textbf{V2v} & \textbf{V2d} & \textbf{V3v} & \textbf{V3d} & \textbf{V4} & \textbf{Ventral} & \textbf{Dorsal} & \textbf{Lateral} \\ \hline
      F (28*28) & 0.8315 & 0.7926 & 0.7795 & 0.7419 & 0.7268 & 0.7323 & 0.7085 & 0.4847 & 0.4831 & 0.4504 \\ 
      S (28*28) & 0.8698 & 0.8340 & 0.8302 & 0.7919 & 0.7808 & 0.7913 & 0.7729 & 0.5796 & 0.6089 & 0.5638 \\ 
      S (4*4) & 0.8432 & 0.8089 & 0.8056 & 0.7690 & 0.7672 & 0.7743 & 0.7425 & 0.5734 & 0.5986 & 0.5513 \\ 
      S (4*4) & 0.7783 & 0.7328 & 0.7374 & 0.6991 & 0.7061 & 0.7043 & 0.7102 & 0.5699 & 0.6002 & 0.5532 \\ \hline
\end{tabular}
\end{center} 
\end{table*}

\subsection{Dependency of Semantic Spatial Transformer Readout on Channel Size}
\label{subsec:stn_channel}

We acknowledge the importance of ensuring that the readout does not skew conclusions about neural representations. The larger improvements for vision models stem from their feature representations having greater spatial dimensions than language models, allowing the SST to better leverage the rich spatial information available in vision models. To mitigate this, we can normalize spatial dimensions across models to ensure uniform treatment. Empirically we show that if we reduce the spatial dimensions of the vision encoder to match those of the language encoder, that does drop the prediction performance and relative gains (Table \ref{tab:stn_channel_analysis}).

The overall trend where higher cortical areas are better modeled by language input and lower cortical areas by visual input is consistently observed across all readouts (Figure. \ref{fig:vision_lang}, \ref{fig:vision_lang_ridge_linear}, \ref{fig:vision_lang_gaussian2D}, \ref{fig:vision_lang_spatial_linear}). However, the margin distinguishing the effectiveness of the models varies slightly. Notably, as we progress from less biologically intuitive readouts to more biologically plausible ones (linear regression, Gaussian 2D, Spatial-Feature Factorized Linear Readout, and finally, the Semantic Spatial Transformer Readout), these trends become increasingly well-defined. Given that the Semantic Spatial Transformer Readout most accurately and consistently models neural responses, we rely on it to delineate regions of the visual cortex sensitive to varying kinds of stimulus information.

\begin{table*}[!ht]
\begin{center} 
\caption{Performance (Analysis of different architectures for Response and Task Optimized models (A - Task Optimized Resnet 50 (pretrained with ImageNet), B - Response Optimized Resnet 50, C - Task Optimized Mask-RCNN (pretrained with MS-COCO), D - Response Optimized Mask-RCNN, E - Response Optimized E2cnn (proposed), all with Semantic-Spatial Transformer Readouts.} 
\label{tab:task_response_architecture} 
\vskip 0.12in
\begin{tabular}{|c|c|c|c|c|c|c|c|c|c|c|}
\hline
\textbf{Encoder Type} & \textbf{V1v} & \textbf{V1d} & \textbf{V2v} & \textbf{V2d} & \textbf{V3v} & \textbf{V3d} & \textbf{V4} & \textbf{Ventral} & \textbf{Dorsal} & \textbf{Lateral} \\ \hline
      A & 0.8507 & 0.8083 & 0.8057 & 0.7603 & 0.7612 & 0.7763 & 0.7674 & 0.6105 & 0.6606 & 0.5823 \\ 
      B & 0.7579 & 0.7034 & 0.7021 & 0.6646 & 0.6861 & 0.6712 & 0.6991 & 0.5546 & 0.5814 & 0.5470 \\ \hline
      C & 0.8543 & 0.8144 & 0.8084 & 0.7693 & 0.7680 & 0.7772 & 0.7793 & 0.6077 & 0.6764 & 0.5987 \\ 
      D & 0.8147 & 0.7654 & 0.7621 & 0.7163 & 0.7089 & 0.6898 & 0.7114 & 0.5648 & 0.5841 & 0.5469 \\ \hline
      E &  0.8698 & 0.8340 & 0.8302 & 0.7919 & 0.7808 & 0.7913 & 0.7729 & 0.5796 & 0.6089 & 0.5638 \\ \hline
\end{tabular}
\end{center} 
\end{table*}

\subsection{Comparing different architectures for Task and Response Optimized models} \label{subsec:task_response_architecture}

Our study carefully controlled several factors to compare task-optimized and response-optimized neural network models for predicting brain responses. Specifically, we held constant both the stimulus set and readout layer, varying only the encoder architecture across models. The rationale for employing different architectures in our study was to leverage state-of-the-art approaches tailored to distinct modeling paradigms. A direct comparison between task-optimized and response-optimized models is inherently challenging due to differences in the available training stimulus sets. Specifically, the stimulus set for training response-optimized models is substantially smaller—approximately 0.03 times the size of the datasets used for task optimization (e.g. ImageNet). Incorporating structural biases into response-optimized models (e.g., rotation equivariance) enables them to learn effectively from smaller datasets. This advantage of rotation-equivariant architectures in neural encoding contexts has been demonstrated in prior studies \cite{khosla2022high} and is a critical factor when designing models that align with the constraints of neural data.

While head-on comparisons using identical architectures for task and neural response optimization could provide valuable insights into the specific contributions of these factors , the primary objective of our study was not to isolate these factors. Instead, we aimed to identify the most predictive models for voxel responses across distinct regions of the visual system. Our findings reveal the current best-performing models for this goal, emphasizing practical predictive utility rather than dissecting the contributions of task versus response optimization in isolation.

We conducted further experiments using - a ResNet-50 encoder trained from scratch exclusively on the NSD dataset, a Mask-RCNN encoder trained from scratch on the NSD dataset, a pretrained Mask-RCNN encoder finetuned on the NSD dataset, and compared it with the proposed task and response optimized encoders in the paper all paired with a Semantic Spatial Transformer readout (Table \ref{tab:task_response_architecture}). We did this to analyze if the same architecture for response- and task-optimized vision models could provide valuable insights. Unlike the task-optimized ResNet-50, which is trained for object classification on ImageNet, the ResNet-50 trained from scratch on neural responses struggled to match the performance of the proposed response-optimized e2cnn model. The task optimized Mask-RCNN model is pretrained on the MS-COCO dataset which is a superset of the images in the NSD dataset. Although both the task optimized performance show a very similar performance, we once again see a similar trend here with the Mask-RCNN encoder trained from scratch on the NSD dataset, where it struggled to reach the performance of the response optimized e2cnn model. This comparison underscores the role of network architecture and the significance of incorporating relevant structural biases into networks when optimizing them on response prediction with limited data (atleast in comparison to large-scale vision datasets).

Task-optimized models, typically pretrained on large-scale datasets (e.g., ImageNet), apply only a linear mapping from their learned representations to brain responses. Although one could examine how diverse architectures and tasks affect performance, prior work \citep{conwell2022large, conwell2024large} suggests that even starkly different architectures (e.g., CNNs vs. transformers) yield similar brain predictivity in task-optimized settings, implying that architecture alone may not be the critical factor. Here, we take a complementary approach by comparing these task-optimized models with response-optimized and LLM-based frameworks, each configured to best align with neural data constraints. Specifically, we select the most effective pretrained architecture for task optimization and pair it with an appropriately chosen architecture for response optimization.

\begin{table*}[!ht]
\begin{center} 
\caption{Performance (Test Accuracies as Normalized Pearson Correlation) of Single Caption Language models - (1) entire sentence is used, (2) Only object words are used, (3) Only stuff words are used, (4) Both object and stuff words are used and (5) Jumbled sentences are used} 
\label{tab:ssingle_caption_analysis} 
\vskip 0.12in
\begin{tabular}{|c|c|c|c|c|c|c|c|c|c|c|}
\hline
\textbf{Caption Type} & \textbf{V1v} & \textbf{V1d} & \textbf{V2v} & \textbf{V2d} & \textbf{V3v} & \textbf{V3d} & \textbf{V4} & \textbf{Ventral} & \textbf{Dorsal} & \textbf{Lateral} \\ \hline
      1 & 0.3974 & \textbf{0.3779} & \textbf{0.3809} & \textbf{0.3702} & \textbf{0.4093} & \textbf{0.4119} & \textbf{0.4882} & \textbf{0.5661} & \textbf{0.6243} & 0.5920 \\ 
      2 & 0.3342 & 0.3252 & 0.3197 & 0.322 & 0.3407 & 0.3506 & 0.4051 & 0.4905 & 0.5287 & 0.5187 \\ 
      3 & 0.3186 & 0.2603 & 0.2888 & 0.2530 & 0.2973 & 0.2868 & 0.3413 & 0.4237 & 0.4418 & 0.4143 \\ 
      4 & 0.3721 & 0.3316 & 0.3439 & 0.3242 & 0.3609 & 0.3575 & 0.4200 & 0.5038 & 0.5415 & 0.5178 \\ 
      5 & \textbf{0.4016} & 0.3759 & 0.3802 & 0.3682 & 0.4080 & 0.4099 & 0.4840 & 0.5615 & 0.6211 & \textbf{0.5952} \\ \hline
\end{tabular}
\end{center} 
\end{table*}

\begin{figure*}[h!]
    \centering
    \includegraphics[width=1.0\linewidth,trim={0 0 0 20},clip]{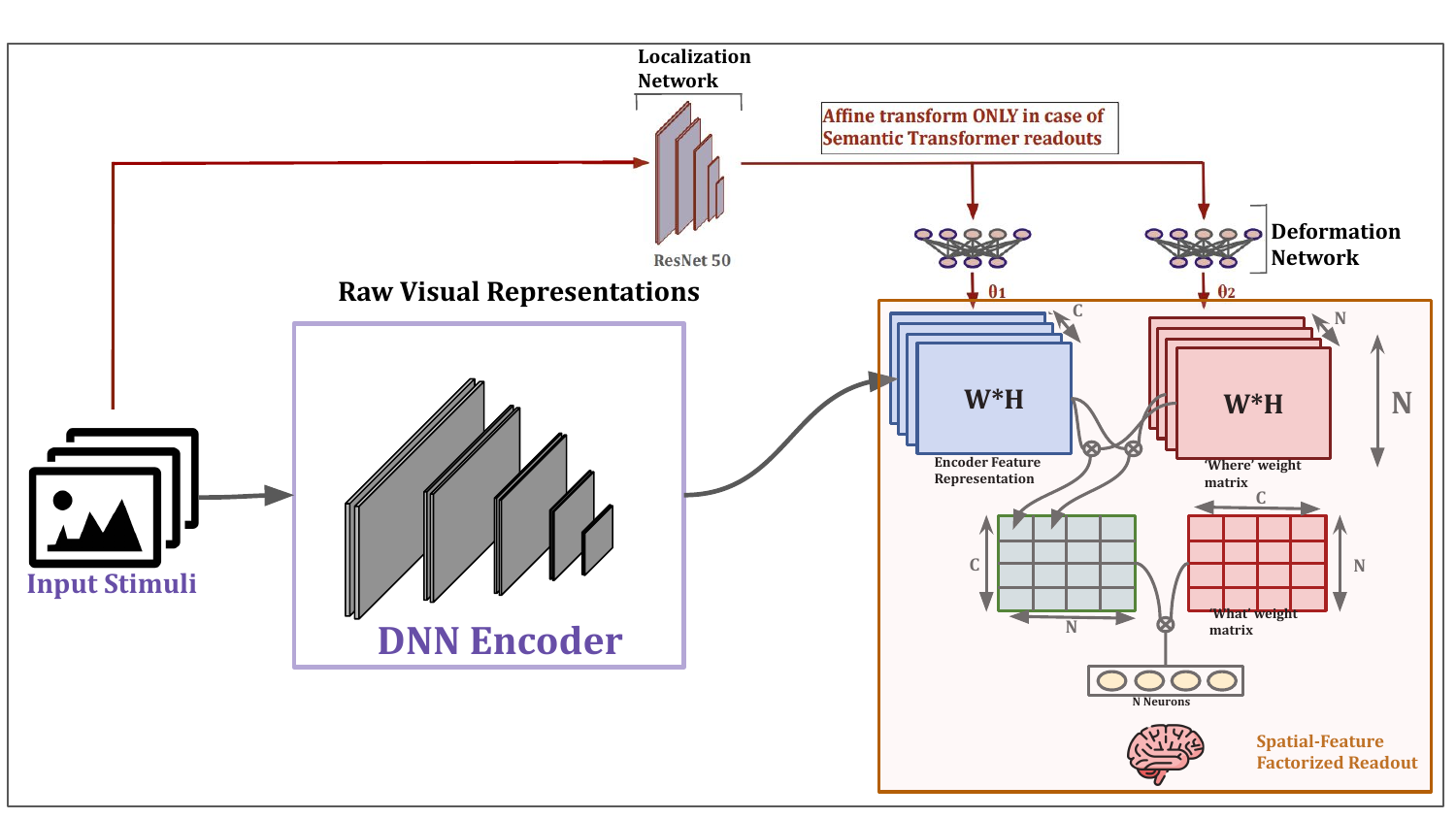}
    \caption{Overall Pipeline when a Semantic Spatial Transformer readout is used.}
    \label{fig:stn_summary}
\end{figure*}

\begin{table*}[!ht]
\begin{center} 
\caption{Number of learnable parameters for each readout configuration - Here, $C$ denotes the number of channels in the encoder feature representation, $N$ is the number of neurons being modeled, and $W×H$ represents the spatial dimensions of each feature map channel.} 
\label{tab:complexity} 
\vskip 0.12in
\begin{tabular}{|c|c|c|c|c|c|c|c|c|c|c|}
\hline
\textbf{Readout Type} & \textbf{Number of parameters learnt} \\ \hline
      Linear & $N*C*H*W$ \\ 
      Gaussian 2D & $N*(C + 7)$ \\ 
      Spatial-Feature Factorized & $N*(C + W*H)$  \\ 
      Semantic Transformer & $N*(C + W*H) + 32*6(N+C) + 196*32*2$ \\ \hline
\end{tabular}
\end{center} 
\end{table*}

\subsection{Further Clarification on the pipeline for Semantic Transformers} 
\label{subsec:stn_details}

Figure \ref{fig:stn_summary} presents an overview of the pipeline when using the Semantic Spatial Transformer Readout. This readout builds upon the existing Spatial-Feature Factorized readout, whose components are highlighted in the orange box in the figure. The key innovation introduced by the Semantic Spatial Transformer is the application of affine transformations to both the encoder feature representation and the spatial weight ("where") matrix, enabling data augmentation and modulation of receptive fields dynamically based on the input. To enable these transformations, the readout incorporates four additional components: (1) Localization Network (2) Deformation Network (seperate for each affine transformation set) (3) Parameterized Sampling Grid (4) Sampler.

The localization network is implemented using a pretrained ResNet-50 block, which generates input stimulus embeddings. Importantly, this network's weights are frozen during training. The motivation for using a pretrained network is to leverage strong, prior-informed embeddings, which can facilitate the learning of effective affine transformations. While the main DNN encoder in the pipeline (whether task-optimized or response-optimized) could also serve as a localization network, we chose a fixed pretrained model to ensure robust and stable representations. Incorporating the main encoder as the localization network is a promising direction for future work.

Each of the two deformation networks is implemented as a linear layer that receives embeddings from the localization network and outputs 6-parameter affine transformations for two distinct purposes - 
\begin{enumerate}
    \item $\theta_1$: transformation parameters for each channel of the encoder feature representation ($R^{C*W*H}$), to apply stimuli dependent data augmentations on each channel.
     \item $\theta_2$: transformation parameters for each neuron in the spatial weight ("where") matrix ($R^{N*W*H}$), that will modulate the respective neuron's receptive field based on the input stimuli.
\end{enumerate}

Once $\theta_1$ and $\theta_2$ are obtained, they are applied to the respective $W \times H$ grids using PyTorch’s built-in affine-grid (to generate sampling grids) and grid-sample (to apply the transformations) functions. The parameterized sampling grid defines how each location in the transformed grid corresponds to coordinates in the original grid. For example, a target coordinate $(x, y)$ in the transformed space might map back to a source coordinate $(i, j)$ in the original grid. Since these source coordinates may not align perfectly with discrete pixel locations, the Sampler uses bilinear interpolation to compute the output value at $(x, y)$ by interpolating values from neighboring pixels around $(i, j)$ in the input.

The affine transformations applied to the encoder feature representations (parameterized by $\theta_1$) for data augmentation purposes are further illustrated in Figure \ref{fig:theta12} A,C,D. Similarly, the transformations applied to the spatial weight matrix (parameterized by $\theta_2$), which allow for dynamic modulation of receptive fields, are detailed in Figure \ref{fig:theta12} B, E.

\paragraph{Computational complexity of the Semantic Spatial Transformer Readout} The Semantic Spatial Transformer introduces minimal overhead—the extra complexity comes solely from two lightweight deformation networks that predict affine transformation parameters for each feature channel in the encoder and one for each voxel. 
The localization network is configured to output embeddings of dimension 196. Each deformation network starts with a linear layer that projects this 196-dimensional embedding to a hidden dimension of 32, which is then further transformed into a 6-parameter affine transformation. The total number of learnable parameters in each deformation network is - 
\begin{enumerate}
    \item For $\theta_1$ (channel-wise transformations): $196*32 + 32*6*C$, where C is thte total number of channels.
    \item For $\theta_2$ (neuron-wise transformations): $196*32 + 32*6*N$, where N is the number of neurons. 
\end{enumerate}

The additional parameters (roughly $32 \cdot 6 \cdot (N + C)$ plus a constant term) are modest relative to the overall parameter count of the encoder. Moreover, the affine grid generation and bilinear sampling operations are computationally efficient and scale linearly with the feature map size. 
Table \ref{tab:complexity} summarizes the number of parameters that need to be learned for each readout configuration.

\textbf{How are dense caption stimuli used with Semantic Transformer readouts?}  
To generate dense caption stimuli, the original image (e.g., of size $424 \times 424$) is first divided into uniform patches of size $8 \times 8$, resulting in a grid of $53 \times 53$ chunks. For each chunk, a caption is generated using a language model (e.g., GPT-2). These captions are then embedded into vector representations using a large language model (LLM). Let the embedding dimension be $M$, which varies depending on the LLM used—for example, $M = 512$ for CLIP, $M = 768$ for MPNET, and $M = 1600$ for GPT-2 XL.  As a result, the dense caption stimuli can be interpreted as an "image" of shape $M \times 53 \times 53$, analogous to a standard RGB image of shape $3 \times 424 \times 424$.

Dense caption stimuli are specifically used in conjunction with a 2-block E2CNN encoder, similar to the response-optimized models used for visual stimuli (which typically use 8 blocks). The output of this encoder is a set of feature maps that can be represented as a $C \times W \times H$ matrix, which integrates naturally with the "what" and "where" matrices in the Semantic Spatial Transformer Readout.  To generate the affine transformations, we do not pass the dense caption stimuli directly. Instead, the original image stimuli are passed through the ResNet-50 localization network to produce more robust and semantically meaningful affine parameters. This design choice is motivated by the desire to leverage strong visual priors from pretrained models.  A promising direction for future work would be to investigate whether affine transformations can be learned directly from linguistic descriptions alone, without relying on the original visual input.

\end{document}